\documentclass[11pt]{article}

\usepackage[utf8]{inputenc} 
\usepackage[T1]{fontenc}    
\usepackage{hyperref}       
\usepackage{url}            
\usepackage{booktabs}       
\usepackage{amsfonts}       
\usepackage{textcomp}

\usepackage{tikz}
\usetikzlibrary{automata,arrows,patterns}
\usepackage{tango-colors}

\usepackage{pgfplots}
\pgfplotsset{compat=1.13}
\usepgfplotslibrary{groupplots}

\usepackage{algorithm}
\usepackage{algpseudocode}

\usepackage{amsmath}
\usepackage{amsthm}
\usepackage{amssymb}
\usepackage{bm}

\usepackage[a4paper, left=3cm, right=2.5cm, top=3cm, bottom=3cm]{geometry}

\usepackage{authblk}
\usepackage{natbib}
\usepackage{doi}

\newtheorem{thm}{Theorem}
\newtheorem{lma}{Lemma}
\theoremstyle{definition}
\newtheorem{dfn}{Definition}

\newcommand{\N}{\mathbb{N}}
\newcommand{\R}{\mathbb{R}}

\newcommand{\rec}{f}
\newcommand{\out}{g}
\newcommand{\cls}{c}
\newcommand{\inps}{\Sigma}
\newcommand{\outps}{\Gamma}
\newcommand{\states}{Q}

\newcommand{\sigCopy}{smooth associative recall}

\newcommand{\IEEEPARstart}[2]{#1#2}

\newcommand{\IEEEpeerreviewmaketitle}{}
\usepackage{changepage}
\renewenvironment{table*}{\begin{table}\begin{adjustwidth}{-2cm}{}}{\end{adjustwidth}\end{table}}

\let\cite\citep

\begin{document}

\title{Reservoir Memory Machines as Neural Computers}

\author[1]{Benjamin Paaßen}
\author[2]{Alexander Schulz}
\author[3]{Terry Stewart}
\author[2]{Barbara Hammer}
\affil[1]{The University of Sydney, Humboldt-University of Berlin}
\affil[2]{Bielefeld University}
\affil[3]{National Research Council of Canada}

\date{\textcopyright 2021 IEEE. Personal use of this material is permitted. Permission from IEEE must be obtained for all other uses, in any current or future media, including reprinting/republishing this material for advertising or promotional purposes, creating new collective works, for resale or redistribution to servers or lists, or reuse of any copyrighted component of this work in other works.
Published article at \citet{Paassen2021TNNLS}} 
\pagestyle{myheadings}
\markright{\textcopyright 2021 IEEE. Published article at \citet{Paassen2021TNNLS}}

\maketitle

\begin{abstract}
Differentiable neural computers extend artificial neural networks with an
explicit memory without interference, thus enabling the model to perform classic computation tasks
such as graph traversal. However, such models are difficult to train, requiring long
training times and large datasets. In this work, we achieve some of the computational
capabilities of differentiable neural computers with a model that can be trained very
efficiently, namely an echo state network with an explicit memory without interference.
This extension enables echo state networks to recognize all regular languages,
including those that contractive echo state networks provably can not recognize.
Further, we demonstrate experimentally that our model performs comparably
to its fully-trained deep version on several typical benchmark tasks for differentiable neural
computers.
\end{abstract}


%
\IEEEpeerreviewmaketitle

\section{Introduction}

\IEEEPARstart{D}{ifferentiable} neural computers (DNCs) are artificial neural networks that
combine a recurrent neural network controller with an external memory to store
information without interference over long stretches of time \cite{NTM}. Such networks have
achieved impressive successes in recent years, solving tasks such as storing inputs
losslessly, performing associative memory recalls, up to question-answering and graph
traversal \cite{DNCMDS,NTM,MANN,Pushdown}. However, DNCs are difficult to train, typically
requiring several ten thousand input sequences until convergence \cite{NTM_impl}.
In this work, we propose a model that can be trained with simple convex optimization
techniques and little data, but still retains a lot of the capabilities of DNCs.
The only sacrifice we need to make for these advantages is that examples of viable memory
access behavior need to be provided as part of the training data.

Architecturally, our proposed model is an echo state network \cite{ESN}
with an explicit external memory to which access is controlled by a convex classifier -
in our case a support vector machine. We show that this external memory
enhances the computational capabilities of echo state networks to strictly more than
finite state machines, whereas contractive echo state networks cannot recognize some
regular languages. More generally, we obtain a model with lossless memory over arbitrarily long
time spans, which extends beyond the abilities of high but finite capacity memory reservoirs in
recent works \cite{MemCap,DeepESN,LMU}.

Finally, our work is an extension of our first version of reservoir memory machines \cite{RMM}.
In particular, we simplify the architecture by merging read and write behavior into a single
classifier, thus making it easier to implement and to train, and we provide a novel model
variant that can solve associative recall tasks, which was beyond the old version. As we
will see in the experiments, a key reason why these changes work is that we now use a
reservoir that can losslessly recall past input states over long stretches of time, namely the
Legendre delay network \cite{LMU}.

In summary, our contributions in this paper are:
\begin{itemize}
\item A novel reservoir neural network architecture - namely the reservoir memory machine (RMM) - which
is equipped with an external memory but can still be trained using convex optimization,
\item a proof that RMMs are strictly more powerful than finite state machines, and
\item a series of experiments, demonstrating that RMMs can solve many benchmark tasks for differentiable
neural computers that are beyond the abilities of standard recurrent models (including deep ones).
\end{itemize}

\section{Background and Related Work}

\subsection{Differentiable Neural Computers}

The original authors define a differentiable neural computer (DNC) as 'a neural network that can
read from and write to an external memory matrix, analogous to the random-access memory in a
conventional computer' \cite{NTM}.
The mechanism to read from and write to memory is typically content-based,
i.e.\ the controller writes to and reads from locations that are similar to a query vector produced
by the controller \cite{NTM}. However, not all memory accesses in computing tasks can be implemented
in a content-based fashion. Therefore, the model is extended with a linking matrix that connects
subsequent write locations in memory and can thus be used during reading to move spatially
in the memory \cite{NTM}. While reviewing the full breadth and depth of neural computing history is
beyond the scope of this paper, we wish to note at least that many variations of this basic setup
exist, such as additional sharpening operations and more effective initialization schemes \cite{DNCMDS,NTM_impl}.
Further, we note the long history of neural computing approaches, going
back at least to the neural pushdown automaton models of \cite{Pushdown}.

Instead, our aim in this work is to develop a network that is as easy to train as possible
while still retaining some of the computational power of DNCs. In particular, we suggest the
following changes.
First, we observe that most benchmark tasks for DNCs can be solved without content-based addressing,
such that our proposed model rather predicts the memory access location directly. We only use
content-based addressing for the associative recall task, where it is necessary.
Second, after writing something to memory, we do not erase it anymore. This enables us to merge the
write and read head: The first access to a memory location is interpreted as writing, all subsequent
accesses as reading.
Third, all our memory accesses are strict and discrete instead of smooth and differentiable.
This reduces memory access to a straightforward classification problem, provided that training data
for memory access behavior is available. We thus also avoid the need for sharpening operations
as suggested in some DNC works \cite{NTM,DNCMDS}.
Finally, and most prominently, we do not train the system end-to-end but only train the memory
address classifier and the mapping from state to output. This enables training via simple convex
optimization, which thus becomes orders of magnitude faster.

While our simplified model can not be expected to achieve the same computational capabilities
as a full DNC (which aims at emulating Turing machines \cite{NTM_impl}), we can show that
our system is strictly more powerful than finite state machines.

\subsection{Finite State Machines}

We analyze the computational power of our system in comparison to finite state machines (FSMs),
in particular \emph{Moore Machines} \cite{FSM}. We define a Moore machine as a $6$-tuple
$(\states, \inps, \outps, \delta, q_0, \rho)$, where $\states$ is a finite set called \emph{states}, $\inps$
is a finite set called \emph{input alphabet}, $\outps$ is a finite set called \emph{output alphabet},
$\delta : \inps \times \states \to \states$ is called the \emph{state transition function}, $q_0 \in \states$ is called the \emph{start state}, and $\rho : \states \to \outps$ is
called the \emph{output function}. A Moore machine
transforms an input sequence $x_1, \ldots, x_T \in \inps^*$ to an output sequence
$z_1, \ldots, z_T \in \outps^*$ via the dynamical system $q_t = \delta(x_t, q_{t-1})$ and
$z_t = \rho(q_t)$ for all $t \in \{1, \ldots, T\}$. We note in passing that Moore machines are a strict
generalization of finite state automata, which can be seen as Moore Machines with the output
alphabet $\outps = \{0, 1\}$ where $\rho(q) = 1$ if $q$ is an accepting state and $\rho(q) = 0$ otherwise.
By virtue of this mechanism, a Moore machine can recognize any regular/Chomsky-3 language \cite{TheoInf}.

\subsection{Relationship of Recurrent Neural Networks and Finite State Machines}
\label{sec:rnns}

Finite state machines are a particularly interesting model for comparison because their dynamics
are very similar to recurrent neural networks. In more detail, we define a \emph{recurrent neural network} 
with $m$ inputs, $n$ neurons, and $K$ outputs
as a $6$-tuple $(\bm{U}, \bm{W}, \vec b, \sigma, \vec h_0, \out)$
of matrices $\bm{U} \in \R^{n \times m}$, $\bm{W} \in \R^{n \times n}$, as well as a bias vector $\vec b \in \R^n$, some function $\sigma : \R \to \R$, an initial state $\vec h_0 \in \R^n$
(typically the zero vector), and a continuous output function $\out : \R^n \to \R^K$.
The system dynamics are defined as follows.
\begin{align}
\vec h_t = &\rec(\vec x_t, h_t) := \sigma\Big(\bm{U} \cdot \vec x_t + \bm{W} \cdot \vec h_{t-1} + \vec b \Big), \label{eq:rnn} \\
\vec y_t = &\out(\vec h_t),  \label{eq:rnn_out}
\end{align}
where $\sigma$ is applied element-wise.

Note that a recurrent neural net has the same Markovian structure as a Moore machine
with $\rec$ being related to $\delta$ and $\out$ to $\rho$. Indeed, it is well known that recurrent
neural network can implement any finite state machine via a correspondence of neurons and
states \cite{Neurautomata}. Even more, by carefully selecting the weights, a recurrent neural
network, even with rational-valued weights, can simulate a full Turing machine \cite{RecTuring}.
Interestingly, this impressive computational power does not hold for contractive reservoir neural
networks, as we will show in the next section.

\subsection{Reservoir Neural Networks and their Computational Limits}

We define a reservoir with $m$ inputs and $n$ neurons as a $5$-tuple
$(\bm{U}, \bm{W}, \vec b, \sigma, \vec h_0)$ with $\bm{U} \in \R^{n \times m}$, $\bm{W} \in
\R^{n \times n}$, $\vec b \in \R^n$, $\sigma : \R \to \R$
and $\vec h_0 \in \R^n$ which has the same dynamics as in Equation~\ref{eq:rnn} but where
$\rec$ ensures the \emph{echo state property}, i.e.\ that the initial state $\vec h_0$ washes out
over time \cite{ESN,EchoStateProperty}.
This property is key to ensure that the state always reacts to the input instead of degenerating to
a stable fix point \cite{ESN} and is typically achieved by designing $\rec$ as a contractive map \cite{MarkovESN}. In particular, we define a contractive reservoir as follows.

\begin{dfn}[Contractive reservoir]
Let $(\bm{U}, \bm{W}, \vec b, \sigma, \vec h_0)$ be a reservoir. We say that this reservoir is
\emph{contractive} with constant $C \in (0, 1)$ if for any two $\vec h, \vec h' \in \R^n$ and any
$\vec x \in \R^m$ it holds:
\begin{equation}
\lVert \rec(\vec x, \vec h) - \rec(\vec x, \vec h') \rVert
\leq C \cdot \lVert \vec h - \vec h' \rVert,
\end{equation}
where $\rec$ is defined as in Equation~\ref{eq:rnn}.
\end{dfn}

For any contractive reservoir it holds that states with the same suffix get arbitrarily close together
for longer suffixes. For example, the states representing the input sequences $b^T$ and $ab^{T-1}$
become indistinguishable for any contractive reservoir for high enough $T$. However, the sequences are
easy to distinguish for a Moore machine (refer to Figure~\ref{fig:moore}), showing that there are
finite state machines that can not be represented by a recurrent neural net based on a contractive 
reservoir, no matter the output function $\out$.

\tikzset{->, >=stealth', semithick}

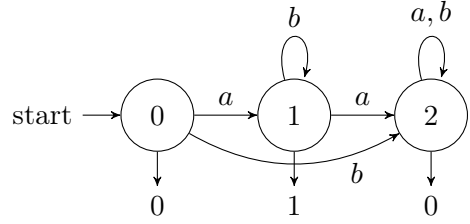
\begin{figure}
\begin{center}
\begin{tikzpicture}
\node[initial, state] (q0) at (0,0) {0};
\node[state] (q1) at (1.8,0) {1};
\node[state] (q2) at (3.6,0) {2};

\node (y0) at (0,-1.2) {0};
\node (y1) at (1.8,-1.2) {1};
\node (y2) at (3.6,-1.2) {0};

\path
(q0) edge (y0)
(q1) edge (y1)
(q2) edge (y2)
(q0) edge[bend right] node[below, pos=0.8] {$b$} (q2)
(q0) edge node[above] {$a$} (q1)
(q1) edge[loop above] node[above] {$b$} (q1)
(q1) edge node[above] {$a$} (q2)
(q2) edge[loop above] node[above] {$a, b$} (q2);

\end{tikzpicture}
\end{center}
\caption{A Moore machine recognizing the language $ab^*$, which can not be recognized by a
contractive echo state network.}
\label{fig:moore}
\end{figure}

\begin{thm}
Consider the Moore Machine $\mathcal{A}_{ab^*}$ illustrated in Figure~\ref{fig:moore}.
Now, let $(\bm{U}, \bm{W}, \vec b, \sigma, \vec h_0)$ be some contractive reservoir
with constant $C \in (0, 1)$. Then, it holds:
For any continuous function $\out : \R^n \to \R$ there exists some $T \in \N$ and
some sequence sequence $x_1, \ldots, x_T \in \{a, b\}^*$ such that
$\out(\vec h_T) \neq z_T$, where $\vec h_T$ is the state of the reservoir for the
one-hot coding sequence of $x_1, \ldots, x_T$, and where $z_T$ is the final
output of the Moore machine $\mathcal{A}_{ab^*}$ for the sequence $x_1, \ldots, x_T$.

\begin{proof}
Let $\vec x_a \in \R^m$ and $\vec x_b \in \R^m$ be the one-hot encodings of the
symbols $a$ and $b$, respectively. 
We define the auxiliary function $\tilde f(\vec h) := \rec(\vec x_b, \vec h)$
with $\rec$ as in Equation~\ref{eq:rnn}.
Because the reservoir is contractive, $\tilde f$ is also contractive.
Hence, the Banach fix point theorem implies that there exists a unique fix point
$\vec h^* \in \R^n$ such that for all $\vec h \in \R^n$:
\begin{equation*}
\lVert \tilde f^T(\vec h) - \vec h^* \rVert \leq C^T \cdot \lVert \vec h - \vec h^* \rVert,
\end{equation*}
where $\tilde f^T$ denotes $T$ applications of $\tilde f$.

Next, any continuous map $\out : \R^n \to \R$ is continuous in $\vec h^*$, specifically.
Accordingly, for any $\epsilon> 0$ there exists some $\delta_\epsilon > 0$, such that for all
$\vec h \in \R^n$ with $\lVert \vec h - \vec h^* \rVert < \delta_\epsilon$ it holds
$|\out(\vec h) - \out(\vec h^*)| < \epsilon$. Let now $\epsilon = \frac{1}{2}$,
let $\vec h_b = \rec(\vec x_b, \vec h_0)$, and let $\vec h_a = \rec(\vec x_a, \vec h_0)$.
Then, there exists some smallest integer $T \in \N$ such that
\begin{equation*}
C^{T-1} \cdot \lVert \vec h_b - \vec h^* \rVert < \delta_\epsilon \text{ and }
C^{T-1} \cdot \lVert \vec h_a - \vec h^* \rVert < \delta_\epsilon .
\end{equation*}

Now, consider the two sequences $b^T$ and $a b^{T-1}$. Note that $\mathcal{A}_{ab^*}$
produces $z_T = 0$ for the first and $z'_T = 1$ for the second sequence.
Further, let $\vec h_T$ and $\vec h'_T$ be the states obtained by the reservoir
via Equation~\ref{eq:rnn} for the first and second sequence,
respectively. Note that we can re-write these states as
$\vec h_T = \tilde f^{T-1}(\vec h_b)$ and $\vec h'_T = \tilde f^{T-1}(\vec h_a)$,
respectively. Therefore, it holds:
\begin{align*}
\lVert \vec h_T - \vec h^* \rVert &=
\lVert \tilde f^{T-1}(\vec h_b) - \vec h^* \rVert
\leq C^{T-1} \cdot \lVert \vec h_b - \vec h^* \rVert < \delta_\epsilon \\
\lVert \vec h'_T - \vec h^* \rVert &=
\lVert \tilde f^{T-1}(\vec h_a) - \vec h^* \rVert
\leq C^{T-1} \cdot \lVert \vec h_a - \vec h^* \rVert < \delta_\epsilon .
\end{align*}
Accordingly, we obtain:
\begin{equation*}
|\out(\vec h_T) - \out(\vec h'_T)| \leq |\out(\vec h_T) - \out(\vec h^*)| + |\out(\vec h'_T) - \out(\vec h^*)| < 2 \epsilon = 1 .
\end{equation*}
However, to ensure $\out(\vec h_T) = z_T = 0$ and $\out(\vec h'_T) = z'_T = 1$, we would
need $|\out(\vec h_T) - \out(\vec h'_T)| = 1$. Since this is not possible,
$\out(\vec h_T) \neq z_T$ or $\out(\vec h'_T) \neq z'_T$ (or both).
\end{proof}
\end{thm}

Importantly, this Theorem only applies if words can be made arbitrarily long. If we consider
any finite language, it is, in principle, possible to construct a contractive reservoir which
can recognize the language.
More generally, \cite{DMM} have shown that the behavior of any contractive recurrent
system can be arbitrarily well approximated in the maximum norm by a definite memory machine,
which is strictly less powerful than a finite state machine. Note that this argument relies on $C < 1$.
It is yet unclear whether reservoirs at the edge of chaos, i.e. $C \geq 1$
\cite{MemCap,DeepESN,MarkovESN,EdgeChaos}, extend this capability.

Despite the limitation in computational power, reservoirs have the advantage that they
can distinguish sequences up to a fixed horizon $T$ very well, without any need to adjust the
matrices $\bm{U}$ or $\bm{W}$ to the data \cite{ESN,MarkovESN,Universality}.
Accordingly, one can leave all reservoir parameters as-is and still obtain a well-performing recurrent
neural net by simply optimizing an output function $\out$, e.g.\ via linear regression \cite{ESN,MarkovESN,Traffic}.
We call such a recurrent neural network an echo state network \cite[ESN]{ESN}.
In this work, we extend an echo state network with an explicit memory from which we can recall past
reservoir states.
Strictly speaking, this violates the echo state property, because we can also recall the initial state.
However, this violation is controlled, because a recall only occurs if a trained classifier says so.
This controlled violation is what raises the computational power beyond finite state machines.

We note that this paper only covers the echo state network perspective on reservoirs.
Liquid state machines \cite{LSM} and the Neural Engineering framework \cite{NEF} provide a
complementary view, using spikes spikes and/or low-rank weight matrices. However, the basic behavior
remains: A reservoir computer is focused on short-term memory and can not distinguish sequences
that have a sufficiently long shared suffix \cite{SpikingCompute}.

Finally, our proposed model is an extension of a prior version of reservoir memory machines
\cite{RMM}, where the model stores past inputs that align well with the output sequence.
In our current work,
we store reservoir states instead of past inputs, such that a single memory entry can summarize a long
stretch of past inputs. To ensure that we do not lose the input information, we also
propose to use Legendre delay networks as reservoirs which support the lossless reconstruction
of inputs from reservoir states \cite{LMU}.
We further replace the alignment mechanism with an additional teaching input
channel which the user can use to specify when data should be stored or read for the training data.
Finally, we simplify the architecture by merging read and write heads into a single
state classifier (except for the associative recall task), thus simplifying training.

\section{Method}

In this section, we describe our main contribution. In particular, we introduce two mechanisms to
extend an echo state network with external memory, prove that this extension suffices to
raise the computational power beyond finite state machines, and provide training algorithms
for both mechanisms.
The first mechanism implements address-based memory with a shared read and write head,
whereas the second mechanism implements associative memory with separate read and write heads.
We use the first mechanism for most of our experimental tasks and the second only for the
associative recall task.

\subsection{Standard Reservoir Memory Machines}

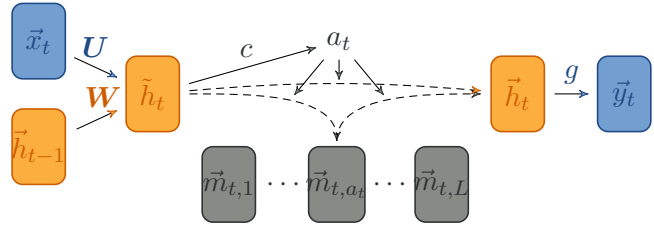
\begin{figure}
\begin{center}
\begin{tikzpicture}
\begin{scope}[shift={(0,+0.7)}]
\draw[fill=skyblue1, draw=skyblue3, semithick, rounded corners] (0,+0.5) rectangle (0.7,-0.5);
\node[skyblue3] (xt) at (0.35,0) {$\vec x_t$};
\end{scope}

\begin{scope}[shift={(0,-0.7)}]
\draw[fill=orange1, draw=orange3, semithick, rounded corners] (0,+0.5) rectangle (0.7,-0.5);
\node[orange3] (hpast) at (0.35,0) {$\vec h_{t-1}$};
\end{scope}

\begin{scope}[shift={(1.5,0)}]
\draw[fill=orange1, draw=orange3, semithick, rounded corners] (0,+0.5) rectangle (0.7,-0.5);
\node[orange3] (tildeht) at (0.35,0) {$\tilde h_t$};
\end{scope}

\path[fill=skyblue1, draw=skyblue3, shorten >=5pt, shorten <=5pt]
(xt) edge node[above, skyblue3] {$\bm{U}$} (tildeht);

\path[fill=orange1, draw=orange3, shorten >=5pt, shorten <=0pt]
(hpast) edge node[above, orange3] {$\bm{W}$} (tildeht);

\begin{scope}[shift={(2.5,-1.2)}]
\draw[fill=aluminium4, draw=aluminium6, semithick, rounded corners] (0,+0.5) rectangle (0.7,-0.5);
\node[aluminium6] (m1) at (0.35,0) {$\vec m_{t,1}$};

\node[aluminium6] at (1.1,0) {$\cdots$};

\end{scope}
\begin{scope}[shift={(3.9,-1.2)}]
\draw[fill=aluminium4, draw=aluminium6, semithick, rounded corners] (0,+0.5) rectangle (0.74,-0.5);
\node[aluminium6] (ml) at (0.37,0) {$\vec m_{t,a_t}$};

\node[aluminium6] at (1.1,0) {$\cdots$};
\end{scope}
\begin{scope}[shift={(5.3,-1.2)}]
\draw[fill=aluminium4, draw=aluminium6, semithick, rounded corners] (0,+0.5) rectangle (0.7,-0.5);
\node[aluminium6] (mL) at (0.35,0) {$\vec m_{t,L}$};
\end{scope}

\node[aluminium6] (at) at (4.3,0.7) {$a_t$};

\begin{scope}[shift={(6.3,0)}]
\draw[fill=orange1, draw=orange3, semithick, rounded corners] (0,+0.5) rectangle (0.7,-0.5);
\node[orange3] (ht) at (0.35,0) {$\vec h_t$};
\end{scope}

\node (write) at (3.6,-0.15) {};
\node (read) at (5,-0.15) {};
\node (copy) at (4.3,0.05) {};

\path[fill=aluminium4, draw=aluminium6]
(tildeht) edge[shorten <=5pt] node[above, aluminium6] {$\cls$} (at)
(tildeht) edge[out=0, in=90, densely dashed, shorten <=5pt, shorten >= 7pt] (ml)
(ml) edge[out=90, in=180, densely dashed, shorten <=7pt, shorten >= 5pt] (ht)
(at) edge (write) edge (read) edge (copy);

\path[draw=orange3,fill=orange1]
(tildeht) edge[bend left=5, densely dashed, shorten <=5pt, shorten >=5pt] (ht);

\begin{scope}[shift={(7.7,0)}]
\draw[fill=skyblue1, draw=skyblue3, semithick, rounded corners] (0,+0.5) rectangle (0.7,-0.5);
\node[skyblue3] (yt) at (0.35,0) {$\vec y_t$};
\end{scope}

\path[fill=skyblue1, draw=skyblue3, shorten >=5pt, shorten <=5pt]
(ht) edge node[above, skyblue3] {$\out$} (yt);

\end{tikzpicture}
\end{center}
\caption{The reservoir memory machine architecture. The preliminary state $\tilde h_t$
is computed via Equation~\ref{eq:rnn} as in standard recurrent nets. From $\tilde h_t$
we compute the memory address $a_t = \cls(\tilde h_t)$. If $a_t = 0$, the memory is ignored
and $\vec h_t = \tilde h_t$. Otherwise, $\tilde h_t$ is written into memory
(if $\vec m_{t, a_t} = \vec 0$) and we set $\vec h_t = \vec m_{t, a_t}$.}
\label{fig:rmm}
\end{figure}

We define a reservoir memory machine (RMM) with $m$ inputs, $n$ neurons, $L$ rows of memory, and $K$ outputs
as an $8$-tuple
$(\bm{U}, \bm{W}, \vec b, \sigma, \vec h_0, \states, \cls, \out)$, where
$(\bm{U}, \bm{W}, \vec b, \sigma, \vec h_0)$ is a reservoir with $m$ inputs and $n$ neurons,
$\states = \{0, \ldots, L\}$ is called the \emph{address set}, $\cls : \R^n \to \states$ is
a classifier, mapping the continuous reservoir state to a memory address,
and $\out : \R^n \to \R^K$ is an output function.
The idea behind our RMM architecture is to maintain the usual
recurrent neural network dynamic as long as $\cls$ outputs the zero address, write to the
memory whenever a nonzero address is selected the first time, and read from memory whenever
a nonzero address is selected a subsequent time.

In more detail, we adjust the system dynamics from Equation~\ref{eq:rnn} as follows,
where $\vec m_{t, l}$ denotes the $l$th memory entry at time $t$.
\begin{align}
\tilde h_t &= \rec(\vec x_t, \vec h_{t-1}) = \sigma\Big(\bm{U} \cdot \vec x_t + \bm{W} \cdot \vec h_{t-1} + \vec b \Big), \notag \\
a_t &= \cls(\tilde h_t) \notag \\
\vec m_{t, l} &= \begin{cases}
\tilde h_t & \text{if } l = a_t \text{ and } \vec m_{t-1, l} = \vec 0 \\
\vec m_{t-1, l} \, & \text{otherwise}
\end{cases} \notag \\
\vec h_t &= \begin{cases}
\tilde h_t & \text{if } a_t = 0 \\
\vec m_{t, a_t} & \text{otherwise}
\end{cases} \label{eq:rmm}
\end{align}
where all memory entries are initialized as $\vec m_{0, l} = \vec 0$, except if $\cls(\vec h_0) > 0$.
In that case, $\vec m_{0, \cls(\vec h_0)} = \vec h_0$.

The output is generated as $\vec y_t = \out(\vec h_t)$ as in Equation~\ref{eq:rnn_out}.

\tikzset{-}

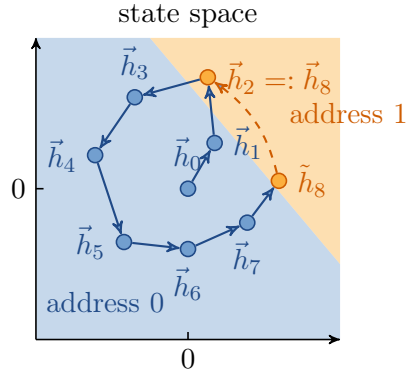
\begin{figure}
\begin{center}
\begin{tikzpicture}
\tikzstyle{point}=[circle, fill=skyblue1, draw=skyblue3, semithick, inner sep=0.07cm]

\draw[fill=skyblue1,draw=none, opacity=0.4] (-2,2) -- (-0.5,2) -- (2,-1) -- (2,-2) -- (-2,-2) -- cycle;
\draw[fill=orange1,draw=none, opacity=0.4] (-0.5,2) -- (2,-1) -- (2,2) -- cycle;

\node[skyblue3, right] at (-2,-1.5) {address 0};
\node[orange3, right]  at (1.2,1) {address 1};

\draw[semithick, ->, >=stealth'] (-2,-2) -- (-2,2);
\draw[semithick, ->, >=stealth'] (-2,-2) -- (2,-2);
\draw[semithick] (-2,0) -- (-1.9,0);
\draw[semithick] (0,-2) -- (0,-1.9);
\node[left] at (-2,0) {$0$};
\node[below] at (0,-2) {$0$};

\node[above] at (0,2) {state space};

\node[point] (h0) at (0,0)     [label={[skyblue3]above:$\vec h_0$}] {};
\node[point] (h1) at (60:0.7)  [label={[skyblue3]right:$\vec h_1$}] {};
\node[point, draw=orange3, fill=orange1] (h2) at (80:1.5)  [label={[orange3]right:{$\vec h_2 =: \vec h_8$}}] {};
\node[point] (h3) at (120:1.4) [label={[skyblue3]above:$\vec h_3$}] {};
\node[point] (h4) at (160:1.3) [label={[skyblue3]left:$\vec h_4$}] {};
\node[point] (h5) at (220:1.1) [label={[skyblue3]left:$\vec h_5$}] {};
\node[point] (h6) at (270:0.8) [label={[skyblue3]below:$\vec h_6$}] {};
\node[point] (h7) at (330:0.9) [label={[skyblue3]below:$\vec h_7$}] {};
\node[point, draw=orange3, fill=orange1] (h8) at (5:1.2)  [label={[orange3]right:$\tilde h_8$}] {};

\path[thick, skyblue3, ->, >=stealth', fill=skyblue1]
(h0) edge (h1)
(h1) edge (h2)
(h2) edge (h3)
(h3) edge (h4)
(h4) edge (h5)
(h5) edge (h6)
(h6) edge (h7)
(h7) edge (h8)
(h8) edge[orange3, dashed, bend right=20] (h2);

\end{tikzpicture}
\end{center}
\caption{An illustration of the reservoir memory machine dynamics. The first time the continuous
state $\vec h_t$ crosses into the receptive field of a memory address $a_t > 0$ (orange
region), it is stored in memory. Whenever the continuous state re-enters this receptive field, the
stored state is recovered (here, for example, $\vec h_8 = \vec h_2$).}
\label{fig:state_space}
\end{figure}

The architecture is illustrated in Figure~\ref{fig:rmm}. The dynamic is further illustrated
in Figure~\ref{fig:state_space}. In particular,
the reservoir memory machine behaves like a regular echo state neural network until the classifier
$\cls$ outputs a nonzero memory address. In that case, we record the current state $\vec h_t$ in memory,
which we recover whenever $\cls$ outputs the same memory address another time (here in time step $8$, where we recover $\vec h_2$).
Note that this is a strict generalization over a standard echo state network
because we recover Equation~\ref{eq:rnn} if $\cls$ is constantly zero.

\subsection{Computational Power}

In this section, we analyze the computational power of our model in more detail.
We first introduce the notion of cycles, both in a Moore machine and in an RMM. Then, we show
that the state of an RMM resulting from any sequence is equivalent to the state resulting from
its cycle-free version. Further, we introduce the notion of a $(\tau, \epsilon)$-distinguishing
reservoir and finally prove our main result, namely that any Moore machine can be implemented
by a reservoir memory machine if the reservoir is $(\tau, \epsilon)$-distinguishing for sufficiently
large $\tau$. We will further show that there exist tasks for which
reservoir memory machines need exponentially less memory compared to finite state machines.

\begin{dfn}[Cycles]\label{dfn:cycles}
Given a Moore machine $\mathcal{A}$ and an input sequence $x_1, \ldots, x_T \in \inps^*$, we define an
\emph{$\mathcal{A}$-cycle} as a subsequence $x_{t'}, \ldots, x_t$ where $0 \leq t' < t \leq T$ and
$q_{t'} = q_t$.

Similarly, given an RMM $\mathcal{M}$ and an input sequence $\vec x_1, \ldots, \vec x_T \in (\R^m)^*$,
we define an \emph{$\mathcal{M}$-cycle} as a subsequence $\vec x_{t'}, \ldots, \vec x_t$ where
$0 \leq t' < t \leq T$ and $a_{t'} = a_t > 0$ (with $a_0 := \cls(\vec h_0)$). Further, we define the
\emph{$\mathcal{M}$-cycle-reduced version} of $\vec x_1, \ldots, \vec x_T$ as the result of the following
recursive procedure: Identify the largest $t$ such that there exists a $t'$
with $\vec x_{t'}, \ldots, \vec x_t$ being an $\mathcal{M}$-cycle. If no such $t$ exists, return
the sequence itself. Otherwise, take the lowest such $t'$ and return the $\mathcal{M}$-cycle-reduced
version of $\vec x_1, \ldots, \vec x_{t'}, \vec x_{t+1}, \ldots \vec x_T$.
\end{dfn}

As an example, consider Figure~\ref{fig:state_space}. There, $\vec x_2, \ldots, \vec x_8$ is an
$\mathcal{M}$-cycle because $a_2 = a_8 = 1 > 0$. The cycle-reduced version of the sequence would
be $\vec x_1, \vec x_2$.

\begin{lma}\label{lma:states}
Let $\mathcal{M}$ be an RMM, let $\vec x_1, \ldots, \vec x_T \in \inps^*$, and let
$\vec x'_1, \ldots, \vec x'_\tau \in \inps^*$ be its $\mathcal{M}$-cycle reduced version.
Then, $\vec h_T = \vec h'_\tau$, i.e.\ the final states for both input sequences are the same.

\begin{proof}
We prove this statement via an induction over the number of cycles in $\vec x_1, \ldots, \vec x_T$.
First, assume that $\vec x_1, \ldots, \vec x_T$ contains no cycles. Then, the claim holds trivially
because the sequence is equal to its cycle-reduced version.

Second, assume that $\vec x_1, \ldots, \vec x_T$ contains at least one cycle and let
$\vec x_{t'}, \ldots, \vec x_t$ be the cycle with largest $t$ and smallest $t'$, i.e.\ the first
cycle that would be removed in Definition~\ref{dfn:cycles}.
Then, consider the sequence $\vec x_1, \ldots, \vec x_{t'}$ and let
$\vec x'_1, \ldots, \vec x'_\tau$ be its cycle-reduced version. Because
$\vec x_1, \ldots, \vec x_{t'}$ contains at least one cycle less than $\vec x_1, \ldots, \vec x_T$,
it follows by induction that $\vec h_{t'} = \vec h'_\tau$. Further, because $a_{t'} = a_t$, we
know that the RMM recalls the same state in $t$ as in $t'$, which yields $\vec h_t = \vec h_{t'}
= \vec h'_\tau$. Finally, we know that $\vec x_{t+1}, \ldots, \vec x_T$ does not add any cycles,
otherwise $t$ would not have been maximal. Therefore, $\vec x'_1, \ldots, \vec x'_\tau, \vec x'_{\tau+1}, \ldots, \vec x'_{\tau+T-t}$ with $\vec x'_{\tau+1}, \ldots, \vec x'_{\tau+T-t}
= \vec x_{t+1}, \ldots, \vec x_T$ is exactly the cycle-reduced version of $\vec x_1, \ldots, \vec x_T$ and we
obtain $\vec h_T = \vec h'_{\tau + T - t}$ as desired. This concludes the proof.
\end{proof}
\end{lma}

This lemma forms one pillar of our main result. The other pillar is a sufficiently rich reservoir,
which we define as follows.

\begin{dfn}[$(\tau, \epsilon)$-distinguishing reservoirs]
Let $\Sigma$ be a subset of $\R^m$.
Further, let $\mathcal{R} = (\bm{U}, \bm{W}, \vec b, \sigma, \vec h_0)$ be a reservoir with $m$ inputs
and $n$ neurons for some $n$. Then, we call $\mathcal{R}$ a $(\tau, \epsilon)$-distinguishing
reservoir on $\Sigma$ if for any two sequences $\vec x_1, \ldots, \vec x_T \in \Sigma^*$
with $T \leq \tau$ and $\vec x'_1, \ldots, \vec x'_{T'} \in \Sigma^*$ with $T' \leq \tau$ and
$\vec x_1, \ldots, \vec x_T \neq \vec x'_1, \ldots, \vec x'_{T'}$ it holds:
$\lVert \vec h_T - \vec h'_{T'}\rVert \geq \epsilon$, where
$\vec h_T$ is the state representing the first and $\vec h'_{T'}$ is the state
representing the second sequence.
\end{dfn}

We note that our notion of $(\tau, \epsilon)$-distinguishing reservoirs is related to the
concept of \emph{memory capacity} \cite{MemCap} in the sense that a memory capacity of $\tau$
requires that input stimuli $\tau$ steps in the past can still be distinguished. Past research
has demonstrated that any finite memory capacity can be achieved with sufficiently many neurons
and a properly constructed reservoir \cite{CRJ2}, such that we assume in the following that a
$(\tau, \epsilon)$-distinguishing reservoir is available for sufficiently distinct input
stimuli $\Sigma$ (like one-hot codings). Importantly, for any finite $\tau$, this property can
be achieved by contractive reservoirs \cite{CRJ2}.

Now follows our main result regarding the computational power of RMMs.

\begin{thm}\label{thm:fsm}
Let $\mathcal{A} = (\states, \inps, \outps, \delta, \rho, q_0)$ be a Moore machine with
$\inps \subset \R^m$, $\states = \{1, \ldots, L\}$, and $\outps = \{1, \ldots, K\}$ for some $m, L, K \in \N$.
Further, let $(\bm{U}, \bm{W}, \vec b, \sigma, \vec h_0)$
be a reservoir with $m$ inputs and $n$ neurons (for some $n \in \N$)
that is $(L, \epsilon)$-distinguishing on $\inps$ for some $\epsilon > 0$.

Then, there exist functions $\cls : \R^n \to \{0, \ldots, L\}$ and $\out : \R^n \to \{1, \ldots, K\}$
such that $\mathcal{M}_\mathcal{A} = (\bm{U}, \bm{W}, \vec b, \sigma, \vec h_0, \{0, \ldots, L\}, \cls, \out)$
is a reservoir memory machine with the following property: For all sequences $\vec x_1, \ldots,
\vec x_T \in \inps^*$ it holds: $a_T = q_T$ and $y_T = z_T$,
where $a_T$ is the memory address of $\mathcal{M}_\mathcal{A}$ at time $T$, $q_T$ is the state
of $\mathcal{A}$ at time $T$, $y_T$ is the output of $\mathcal{M}_\mathcal{A}$ at time $T$, and
$z_T$ is the output of $\mathcal{A}$ at time $T$.
\begin{proof}
First, we introduce two auxiliary functions which compute the state of $\mathcal{A}$ and
$\mathcal{M}_\mathcal{A}$ respectively, in particular: $\Delta(\vec x_1, \ldots, \vec x_T) := \delta(\vec x_T, \Delta(\vec x_1, \ldots, \vec x_{T-1}))$
with $\Delta(\varepsilon) = q_0$ and $F(\vec x_1, \ldots, \vec x_T) = \sigma(\bm{U} \cdot \vec x_T + \bm{W}
\cdot F(\vec x_1, \ldots, \vec x_{T-1}) + \vec b)$ with $F(\varepsilon) = \vec h_0$.

Next, we define two auxiliary sets, namely a) the set $\mathcal{X}_0$ of all sequences
$\vec x_1, \ldots, \vec x_T$ which contain no $\mathcal{A}$-cycle,
and b) the set $\mathcal{X}_1$ of all sequences $\vec x_1, \ldots, \vec x_T$
where $\vec x_1, \ldots, \vec x_{T-1}$ contains no $\mathcal{A}$-cycle
but there exists some $t < T$ such that $\Delta(\vec x_1, \ldots, \vec x_t) = \Delta(\vec x_1, \ldots, \vec x_T)$, i.e.\ the sequence
contains exactly one cycle that ends in $T$. Note that $T \leq |\states| = L$, otherwise
$\vec x_1, \ldots, \vec x_{T-1}$ would contain a cycle.

Now, consider the set of tuples $\mathcal{H} = \{\big(F(\bar x), \Delta(\bar x)\big) |
\bar x \in \mathcal{X}_0 \cup \mathcal{X}_1\}$. Because we required that the
reservoir $(\bm{U}, \bm{W}, \vec b, \sigma, \vec h_0)$
is $(L, \epsilon)$-distinguishing, we obtain for any two
$(\vec h, q), (\vec h', q') \in \mathcal{H}$: If $\lVert \vec h - \vec h' \rVert < \epsilon$,
then $q = q'$. Accordingly, we can construct $\cls$ as a one-nearest neighbor classifier and obtain
$\cls(F(\bar x)) = \Delta(\bar x)$ for all $\bar x \in \mathcal{X}_0 \cup
\mathcal{X}_1$.

We next show via induction over the number of $\mathcal{A}$-cycles that this classifier suffices
to ensure $a_T = q_T$ for \emph{any} input sequence.
First let $\vec x_1, \ldots, \vec x_T \in \inps^*$ be $\mathcal{A}$-cycle free. Then,
$a_T = \cls(F(\vec x_1, \ldots, \vec x_T)) = \Delta(\vec x_1, \ldots, \vec x_T) = q_T$
follows immediately from our classifier construction above. Now, assume that
$\vec x_1, \ldots, \vec x_T$ contains at least one $\mathcal{A}$-cycle and let $\vec x_r, \ldots, \vec x_s$ with
$r < s$ be the $\mathcal{A}$-cycle with largest $s$ and smallest $r$.
Because $\vec x_1, \ldots, \vec x_{s-1}$ contains at least one $\mathcal{A}$-cycle less than before,
we know by induction that $q_t = a_t$ for $t < s$. Accordingly, the equality also holds for
the $\mathcal{M}$-cycle reduced version $\vec x'_1, \ldots, \vec x'_\tau$ of $\vec x_1, \ldots, \vec x_{s-1}$,
which hence implies that $\vec x'_1, \ldots, \vec x'_\tau$ is also $\mathcal{A}$-cycle free.
Now, consider the sequence $\vec x'_1, \ldots, \vec x'_\tau, \vec x_s$. Since $\vec x'_1, \ldots, \vec x'_\tau$
is $\mathcal{A}$-cycle free, $\vec x'_1, \ldots, \vec x'_\tau, \vec x_s$ must lie either in 
$\mathcal{X}_0$ or $\mathcal{X}_1$. Accordingly, our classifier construction
ensures that $q_s = q'_{\tau+1} = a'_{\tau+1}$. Further, thanks to Lemma~\ref{lma:states}, we know
that the state $\vec h'_\tau$ is equal to $\vec h_{s-1}$, which in turn implies that
$a'_{\tau+1} = a_s$.
Next, because $\vec x_{s+1}, \ldots, \vec x_T$ adds no $\mathcal{A}$-cycles
(otherwise $s$ would not have been maximal), $\vec x'_1, \ldots, \vec x'_\tau, \vec x_{s+1}, \ldots, \vec x_T$ is
$\mathcal{A}$-cycle free, which means that our classifier construction ensures
that the states $q_{s+t} = q'_{\tau+t} = a'_{\tau+t}$ for all $t \in \{1, \ldots, T-s\}$.
Lemma~\ref{lma:states} then yields $a'_{\tau+t} = a_{s+t}$, which concludes our proof by induction.

It remains to show that $y_T = z_T$. Consider the set of tuples
$\mathcal{Y} = \{ \big(F(\bar x), \rho(\Delta(\bar x))\big) | \bar x \in \mathcal{X}_0 \}$.
With the same reasoning as before, this yields a well-defined training data set for a
$1$-nearest neighbor classifier $\out$, which ensures $\out(F(\bar x)) = \rho(\Delta(\bar x))$
for all $\bar x \in \mathcal{X}_0$.

Furthermore, this construction suffices to ensure $y_T = z_T$ for \emph{any} input sequence
$\vec x_1, \ldots, \vec x_T$. In particular, Lemma~\ref{lma:states} guarantees that
$\vec h_T = F(x'_1, \ldots, x'_\tau)$ where $\vec x'_1, \ldots, \vec x'_\tau$
is the $\mathcal{M}$-cycle free version of $\vec x_1, \ldots, \vec x_T$.
Further, because $q'_t = a'_t$ for all $t \in \{1, \ldots, \tau\}$,
$\vec x'_1, \ldots, \vec x'_\tau$ is also $\mathcal{A}$-cycle free, which
ensures that $x'_1, \ldots, x'_\tau \in \mathcal{X}_0$ and, hence,
$y_T = \out(\vec h_T) = \out(F(\vec x'_1, \ldots, \vec x'_\tau))
= \rho(\Delta(\vec x'_1, \ldots, \vec x'_\tau)) = \rho(\Delta(\vec x_1, \ldots, \vec x_T))
= z_T$. This concludes the proof.
\end{proof}
\end{thm}

We note in passing that a one-nearest neighbor classifier is only one possible implementation
of $\cls$ and $\out$. In practice, we use support vector machines (SVMs).
SVMs fit well because they are maximum margin classifiers and, as such, can exploit
the $(\tau, \epsilon)$-distinguishing property of reservoirs \cite{RBFSVM}.

We have shown that RMMs are at least as powerful as finite state machines. We now proceed
to show that there is at least one task which an RMM can solve
for which a finite state machine would require exponentially or infinitely many states.

\begin{thm}\label{thm:copy}
Let $\inps \subset \R^m$ be a finite set with $\vec 0, \vec 1 \in \inps$. We define the $(\inps, \tau)$-copy
task for some $\tau \in \N$ as follows:
For any input sequence $\vec x_1, \ldots, \vec x_T, \vec 1, \vec 0, \ldots, \vec 0
\in \inps^*$ with $\vec x_1, \ldots, \vec x_T \notin \{\vec 0, \vec 1\}^*$, a suffix of $T-1$ zeros,
and $T \leq \tau$ we define the desired output sequence
$\vec x_1, \ldots, \vec x_T, \vec x_1, \ldots, \vec x_T$, i.e.\ the input is copied once.

a) Let $(\bm{U}, \bm{W}, \vec b, \sigma, \vec h_0)$ be a reservoir with $m$ inputs and $n$
neurons that is $(\tau+1, \epsilon)$-distinguishing on $\inps$ for some $\epsilon > 0$.
Then, there exist two functions $\cls : \R^n \to \{1, \ldots, \tau\}$ and $\out : \R^n \to \inps$
such that the reservoir memory machine $(\bm{U}, \bm{W}, \vec b, \sigma, \vec h_0,
\{0, \ldots, \tau\}, \cls, \out)$ solves the copy task.

b) Any Moore machine that solves the copy task has at least $|\inps|^T$ states
\begin{proof}
a) Because the reservoir is $(\tau+1, \epsilon)$-distinguishing, the sets
$\mathcal{H}_T = \{ \vec h_T \}$ containing states resulting from sequences of length $T \leq \tau$
are distinguishable with a margin of at least $\epsilon$. Accordingly, we can construct
a classifier $\cls : \R^n \to \{1, \ldots, \tau\}$ (e.g.\ a one-nearest neighbor-classifier) that maps 
$\cls(\vec h_T) = T$ for all $T \in \{1, \ldots, \tau\}$. Further, because the reservoir
is $(\tau+1, \epsilon)$-distinguishing, we can also ensure that the classifier maps
$\cls(\vec h_{T+1}) = 1$ if $\vec h_{T+1}$ is a state representing a sequence
$\vec x_1, \ldots, \vec x_T, \vec 1$ with $T \leq \tau$. Finally, also by the
$(\tau+1, \epsilon)$-distinguishing property, we can construct a classifier
$\out : \R^n \to \inps$ which maps the state representing each sequence up to length $T$
to its last symbol, i.e.\ for all $\vec x_1, \ldots, \vec x_T \in \inps^*$ with $T \leq \tau$,
 we obtain $\out(\vec h_T) = \vec x_T$.

Now, let $\vec x_1, \ldots, \vec x_T, \vec 1, \vec 0, \ldots, \vec 0
\in \inps^*$ with $T \leq \tau$ be an input sequence for the $(\inps, \tau)$-copy task.
By construction, we obtain the memory
addresses $a_t = \cls(\tilde h_t) = t$ for $t \in \{1, \ldots, T\}$ and $a_{T+1} = 1$
because the sequence $\vec x_1, \ldots, \vec x_T, \vec 1$ ends in $\vec 1$. Accordingly, the RMM
re-sets the state to $\vec h_{T+1} = \vec h_1$ and, hence, the preliminary state $\tilde h_{T+2}$
corresponds to the input sequence $\vec x_1, \vec 0$, which in turn results in
$\cls(\tilde h_{T+2}) = 2$, such that we recover $\vec h_{T+2} = \vec h_2$ and so forth,
i.e.\ the state sequence is $\vec h_1, \ldots, \vec h_T, \vec h_1, \ldots, \vec h_T$.
Further, $\out(\vec h_t) = \vec x_t$, which implies that the RMM solves the copy task.

b) Assume there exists a Moore machine with less than $|\inps|^T$ states that solves the copy task.
Because there are $|\inps|^T$ different sequences of length $T$ over $\inps$, there must thus
exist two sequences $\vec x_1, \ldots, \vec x_T \neq \vec x'_1, \ldots, \vec x'_T$ over $\inps$ which lead to the
same state. Accordingly, if we now input $T$ zeros, the output of the Moore Machine
will be the same as well, which is incorrect for at least one of the two sequences.
\end{proof}
\end{thm}

We note that, for fixed sequence length $T$, this task is simple to solve for a regular echo state
network by training $\out$ to return $\out(\vec h_t) = \vec x_t + \vec x_{t-T}$.
However, we define the copy task with variable sequence length, which is not obviously
solvable for an ESN but remains simple for an RMM. Further, by extending $\Sigma$ to an infinite set
(e.g. $\R^m$), the same construction yields that no Moore machine can solve the copy task, whereas
an RMM still can (as we show empirically in the experiments).

\subsection{Associative Memory}

To implement associative memory, we separate write and read access to the memory.
Roughly speaking, we change $\cls$ to a binary classifier that only decides whether we want
to write to memory or not, and we read from memory whenever a) we don't write to it and b) the
current state is similar enough to some state stored in memory.
More precisely, we define an associative RMM (aRMM) with $m$ inputs, $n$ neurons, $L$ rows of memory,
$d$ latent dimensions and $K$ outputs as an $11$-tuple
$(\bm{U}, \bm{W}, \vec b, \sigma, \vec h_0, Q, \cls, \phi, \psi, \theta, \out)$ where
$(\bm{U}, \bm{W}, \vec b, \sigma, \vec h_0)$ is a reservoir with $m$ inputs and $n$ neurons as before,
$Q = \{1, \ldots, L\}$ is an address set as before, $\out : \R^n \to \R^K$ is an output function
as before, but $\cls : \R^n \to \{0, 1\}$ is now a binary classifier called write head,
$\theta \in \R^+$ is a threshold, and $\phi : \R^n \to \R^d$ as well as
$\psi : \R^n \to \R^d$ are auxiliary mappings into some latent space in which we measure distance for
the purpose of association. In particular, the aRMM dynamic is:
\begin{align}
\tilde h_t &= \rec(\vec x_t, \vec h_{t-1}) = \sigma\Big(\bm{U} \cdot \vec x_t + \bm{W} \cdot \vec h_{t-1} + \vec b \Big), \notag \\
a_t &= a_{t-1} + \cls(\tilde h_t) \notag \\
\vec m_{t, l} &= \begin{cases}
\tilde h_t & \text{if } l = a_t \text{ and } \cls(\tilde h_t) = 1 \\
\vec m_{t-1, l} \, & \text{otherwise}
\end{cases} \notag \\
\vec h_t &= \begin{cases}
\tilde h_t & \text{if } \cls(\tilde h_t) = 1 \text{ or } \forall l: d_{t, l}^2 > \theta  \\
\vec m_{t, l^*} & \text{otherwise, with } l^* = \arg\min_l d_{t, l}^2
\end{cases}\notag \\
d_{t, l}^2 &= \lVert \phi(\tilde h_t) - \psi(\vec m_{t, l}) \rVert^2  \label{eq:rmm_assoc}
\end{align}
where all $\vec m_{0, l} = \vec 0$ and $a_0 = 0$. Note that, once the memory is full,
additional writes are ignored.

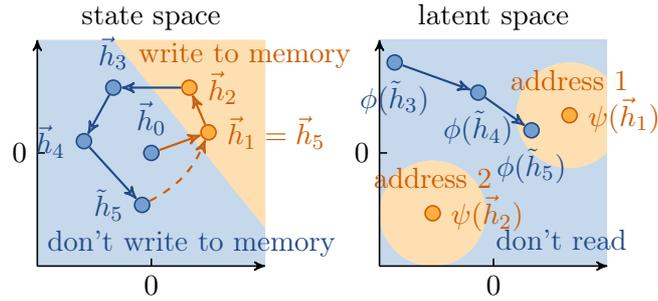
\begin{figure}
\begin{center}
\begin{tikzpicture}
\tikzstyle{point}=[circle, fill=skyblue1, draw=skyblue3, semithick, inner sep=0.07cm]

\begin{scope}[shift={(-2,0)}]

\node[above] at (0,1.5) {state space};

\draw[fill=skyblue1,draw=none, opacity=0.4] (-1.5,1.5) -- (-0.5,1.5) -- (1.5,-1) -- (1.5,-1.5) -- (-1.5,-1.5) -- cycle;
\draw[fill=orange1,draw=none, opacity=0.4] (-0.5,1.5) -- (1.5,-1) -- (1.5,1.5) -- cycle;

\node[skyblue3, right] at (-1.5,-1.2) {don't write to memory};
\node[orange3, right]  at (-0.3,1.3) {write to memory};

\draw[semithick, ->, >=stealth'] (-1.5,-1.5) -- (-1.5,1.5);
\draw[semithick, ->, >=stealth'] (-1.5,-1.5) -- (1.5,-1.5);
\draw[semithick] (-1.5,0) -- (-1.4,0);
\draw[semithick] (0,-1.5) -- (0,-1.4);
\node[left] at (-1.5,0) {$0$};
\node[below] at (0,-1.5) {$0$};

\node[point] (h0) at (0,0) [label={[skyblue3]above:$\vec h_0$}] {};
\node[point, draw=orange3, fill=orange1] (h1) at (20:0.8) [label={[orange3]right:{$\vec h_1 = \vec h_5$}}] {};
\node[point, draw=orange3, fill=orange1] (h2) at (60:1)  [label={[orange3]right:{$\vec h_2$}}] {};
\node[point] (h3) at (120:1) [label={[skyblue3]above:$\vec h_3$}] {};
\node[point] (h4) at (170:0.9) [label={[skyblue3]left:$\vec h_4$}] {};
\node[point] (h5) at (260:0.7) [label={[skyblue3]left:$\tilde h_5$}] {};

\path[thick, skyblue3, ->, >=stealth', fill=skyblue1]
(h0) edge[orange3] (h1)
(h1) edge[orange3] (h2)
(h2) edge (h3)
(h3) edge (h4)
(h4) edge (h5)
(h5) edge[orange3, dashed, bend right=20] (h1);
\end{scope}

\begin{scope}[shift={(+2.5,0)}]

\node[above] at (0,1.5) {latent space};

\draw[fill=skyblue1,draw=none, opacity=0.4] (-1.5,-1.5) rectangle (1.5,1.5);

\draw[fill=white, draw=none] (1,0.5) circle (0.7);
\draw[fill=orange1, opacity=0.4, draw=none] (1,0.5) circle (0.7);
\draw[fill=white, draw=none] (-0.8,-0.8) circle (0.7);
\draw[fill=orange1, opacity=0.4, draw=none] (-0.8,-0.8) circle (0.7);
\draw[fill=white, draw=none] (1.5,1.5) rectangle (1.8,-1.5);

\node[skyblue3, right] at (-0.1,-1.15) {don't read};
\node[orange3, above=0.2cm] at (1,0.5) {address 1};
\node[orange3, above=0.2cm] at (-0.8,-0.8) {address 2};

\draw[semithick, ->, >=stealth'] (-1.5,-1.5) -- (-1.5,1.5);
\draw[semithick, ->, >=stealth'] (-1.5,-1.5) -- (1.5,-1.5);
\draw[semithick] (-1.5,0) -- (-1.4,0);
\draw[semithick] (0,-1.5) -- (0,-1.4);
\node[left]  at (-1.5,0) {$0$};
\node[below] at (0,-1.5) {$0$};

\node[point, draw=orange3, fill=orange1] (h1) at (1,0.5) [label={[orange3]right:{$\psi(\vec h_1)$}}] {};
\node[point, draw=orange3, fill=orange1] (h2) at (-0.8,-0.8)  [label={[orange3]right:{$\psi(\vec h_2)$}}] {};
\node[point] (h3) at (-1.3,1.2) [label={[skyblue3]below:$\phi(\tilde h_3)$}] {};
\node[point] (h4) at (-0.2,0.8) [label={[skyblue3]below:$\phi(\tilde h_4)$}] {};
\node[point] (h5) at (0.5,0.3) [label={[skyblue3]below:$\phi(\tilde h_5)$}] {};

\path[thick, skyblue3, ->, >=stealth', fill=skyblue1]
(h3) edge (h4)
(h4) edge (h5);
\end{scope}

\end{tikzpicture}
\end{center}
\caption{An illustration of the aRMM dynamics.
Left: Any state in the orange region is written to memory. The stored states are then mapped
via $\psi$ to a latent space (right) and compared with the current latent state
$\phi(\vec h_t)$. Whenever $\phi(\vec h_t)$ enters a $\theta$-ball around $\psi(\vec m)$
for some memory state $\vec m$, $\vec m$ is read from memory and overrides $\vec h_t$.}
\label{fig:state_space_assoc}
\end{figure}

The system dynamic is illustrated in Figure~\ref{fig:state_space_assoc}.
The orange region in the left part of the figure corresponds to the receptive field of the write head, i.e.\ where
$\cls(\tilde h_t) = 1$. All states in that region are written to memory until the memory is full.
The association mechanism works as follows. We map the current state $\tilde h_t$ to
a latent space (right part of the figure) via the mapping $\phi$ and all memory states to the same
space via a mapping $\psi$.
Whenever $\phi(\tilde h_t)$ enters a $\theta$-ball (orange regions) around some memory state
$\psi(\vec m_{t, l})$, we set $\vec h_t = \vec m_{t, l}$. If multiple $\theta$-balls are entered,
we take the closest memory state.

\subsection{Training Reservoir Memory Machines}

The first step in setting up a reservoir memory machine is to initialize a reservoir
$(\bm{U}, \bm{W}, \vec b, \sigma, \vec h_0)$ that is expressive enough to enable address
classification as well as the output mapping.
Although our approach is agnostic regarding the choice of reservoir, we generally recommend Legendre
delay networks as reservoirs because they are designed to losslessly 
extract past states via linear operations, have few hyperparameters, and are deterministically
constructed, which avoids issues of unfortunate random initializations \cite{LMU}. We will also see
that this type of reservoir performs best on our benchmarks.

\begin{algorithm}
\caption{The training algorithm for a reservoir memory machine
given a reservoir $(\bm{U}, \bm{W}, \vec b, \sigma, \vec h_0)$
with $m$ inputs and $n$ neurons,
as well as training inputs $\vec x_1, \ldots, \vec x_T$, training addresses $a_1, \ldots, a_T$,
and training outputs $\vec y_1, \ldots, \vec y_T$. For multiple training sequences,
lines 3-14 need to be repeated.}
\label{alg:train_rmm}
\begin{algorithmic}[1]
\Function{train\_rmm}{Reservoir $(\bm{U}, \bm{W}, \vec b, \sigma, \vec h_0)$,
training data $\vec x_1, \ldots, \vec x_T$, $a_1, \ldots, a_T$,
and $\vec y_1, \ldots, \vec y_T$}
\State $L \gets \max \{a_1, \ldots, a_T\}$.
\State Initialize $\bm{M}$ as an $L \times n$ matrix.
\For{$t \gets 1, \ldots, T$}
\State $\tilde h_t \gets \sigma\Big(\bm{U} \cdot \vec x_t + \bm{W} \cdot \vec h_{t-1} + \vec b \Big)$.
\If{$a_t > 0$}
\If{$\vec m_{a_t} = \vec 0$}
\State $\vec m_{a_t} \gets \tilde h_t$.
\EndIf
\State $\vec h_t \gets \vec m_{a_t}$.
\Else
\State $\vec h_t \gets \tilde h_t$.
\EndIf
\EndFor
\State Train function $\out$ via linear regression with training data
$\big\{(\vec h_t, \vec y_t) | t \in \{1, \ldots, T\}\big\}$.
\State Train classifier $\cls$ (e.g.\ an SVM) with
training data
$\big\{(\tilde h_t, a_t) | t \in \{1, \ldots, T\}\big\}$.
\State \Return $(\bm{U}, \bm{W}, \vec b, \sigma, \vec h_0, \{0, \ldots, L\}, \cls, \out)$.
\EndFunction
\end{algorithmic}
\end{algorithm}

Once a reservoir is set up, we require training data in the form of three sequences
$\vec x_1, \ldots, \vec x_T \in (\R^m)^*$, $a_1, \ldots, a_T \in \{0, \ldots, L\}^*$, and
$\vec y_1, \ldots, \vec y_T \in (\R^K)^*$, where $\vec x_1, \ldots, \vec x_T$ are the inputs,
$a_1, \ldots, a_T$ are the desired memory addresses, and $\vec y_1, \ldots, \vec y_T$ are the
desired outputs. The fact that the memory addresses are part of the training data makes our approach
less autonomous compared to differentiable neural computers, which can learn the memory addresses \cite{NTM}. However, we would argue that the memory access patterns for the training data are
typically straightforward to construct, at least for all tasks in our experiments (see there).
If this is not the case, one would develop data-driven heuristics to recognize
special states which need to be stored and recovered, such as a clustering based on the output
or an alignment of input and output states as recommended in our past work \cite{RMM}.

Once such training data is constructed, we can compute the state sequence $\vec h_1, \ldots, \vec h_T$
by applying the dynamics in Equation~\ref{eq:rmm} or~\ref{eq:rmm_assoc}, where $a_t$ is given by the
training data instead of a classifier, i.e.\ we perform teacher forcing. We can then train the
output function $\out$ via linear regression or any other fast optimization, using the
training data pairs $(\vec h_t, \vec y_t)$.

The final step of our training mechanism is the state classifier $\cls$. If we wish to train a standard
RMM, we can directly use the pairs $(\tilde h_t, a_t)$ as training data. Again, our approach is agnostic
to the choice of classifier but we use support vector machines in practice 
because they are swift to train and, by virtue of being maximum margin classifiers,
can exploit the $(\tau, \epsilon)$-distinguishing property of
reservoirs. This yields the final RMM
$(\bm{U}, \bm{W}, \vec b, \sigma, \vec h_0, \{0, \ldots, L\}, \cls, \out)$.
The exact training scheme is shown in Algorithm~\ref{alg:train_rmm}.

\begin{algorithm}
\caption{The training algorithm for an associative reservoir memory machine
given a reservoir $(\bm{U}, \bm{W}, \vec b, \sigma, \vec h_0)$
with $m$ inputs and $n$ neurons,
as well as training inputs $\vec x_1, \ldots, \vec x_T$, training addresses $a_1, \ldots, a_T$,
and training outputs $\vec y_1, \ldots, \vec y_T$.}
\label{alg:train_armm}
\begin{algorithmic}[1]
\Function{train\_armm}{Reservoir $(\bm{U}, \bm{W}, \vec b, \sigma, \vec h_0)$,
training data $\vec x_1, \ldots, \vec x_T$, $a_1, \ldots, a_T$,
and $\vec y_1, \ldots, \vec y_T$}
\State $L \gets \max \{a_1, \ldots, a_T\}$.
\State Initialize $\bm{M}$ as an $L \times n$ matrix.
\State Initialize $\mathcal{H}$ as empty set.
\For{$t \gets 1, \ldots, T$}
\State $\tilde h_t \gets \sigma\Big(\bm{U} \cdot \vec x_t + \bm{W} \cdot \vec h_{t-1} + \vec b \Big)$.
\State $\hat a_t \gets 0$.
\If{$a_t > 0$ and $\vec m_{a_t} = \vec 0$}
\State $\vec m_{a_t} \gets \tilde h_t$. \Comment{Write to memory}
\State $\hat a_t \gets 1$.
\Else
\For{$l \in \{1, \ldots, L\} \setminus \{a_t\}$}
\State Add $(\tilde h_t, \vec m_l, -1)$ to $\mathcal{H}$.
\EndFor
\If{$a_t > 0$}
\State Add $(\tilde h_t, \vec m_{a_t}, +1)$ to $\mathcal{H}$.
\State $\vec h_t \gets \vec m_{a_t}$. \Comment{Read from memory}
\Else
\State $\vec h_t \gets \tilde h_t$.
\EndIf
\EndIf
\EndFor
\State Train function $\out$ via linear regression with training data
$\big\{(\vec h_t, \vec y_t) | t \in \{1, \ldots, T\}\big\}$.
\State Train classifier $\cls$ (e.g.\ an SVM) with
training data
$\big\{(\tilde h_t, \hat a_t) | t \in \{1, \ldots, T\}\big\}$.
\State Train $\phi$, $\psi$, and $\theta$ via problem~\ref{eq:assoc_lp} for
training data $\mathcal{H}$.
\State \Return $(\bm{U}, \bm{W}, \vec b, \sigma, \vec h_0, Q, \cls, \phi, \psi, \theta, \out)$.
\EndFunction
\end{algorithmic}
\end{algorithm}

By contrast, training an associative RMM is more complicated.
We begin with the write head $\cls$. In particular, we create a new address sequence
$\hat a_1, \ldots, \hat a_T$ where $\hat a_t = 1$ if $a_t > 0$ and there is no $t' < t$ with
$a_t = a_{t'}$, and $\hat a_t = 0$ otherwise. Then, we train the write head $\cls$ as a 
binary classifier (e.g.\ a Gauss-kernel SVM) on the training data $(\tilde h_t, \hat a_t)$.
Now, the read head remains, i.e.\ we need to find $\phi$ and $\psi$ as well as $\theta$, such that
$\lVert \phi(\tilde h_t) - \psi(\vec m_{t, l}) \rVert^2$ smaller $\theta$ if state $\tilde h_t$ and
memory line $\vec m_{t, l}$ should be associated and larger or equal to $\theta$ otherwise.
For this purpose, we re-write the data as triples $(\tilde h_i, \vec m_i, z_i)$ with $z_i = +1$ if
$\tilde h_i$ and $\vec m_i$ should be associated and $z_i = -1$ otherwise.
This can be seen as a metric learning problem \cite{MetricLearn} or a transfer learning problem
\cite{TransferLearn}, either of which is hard to solve in general. Fortunately, our specific problem
is simpler because any state $\tilde h_t$ can only encode the past inputs. Accordingly, we
construct linear operators $\bm{\Phi}_1, \ldots, \bm{\Phi}_\tau \in (\R^{m \times n})^*$ which
reconstruct the input $t$ steps before the current state (for example by using Legendre delay units
\cite{LMU}). Then, we solve the problem:
\begin{align}
\min_{\bm{A} \in \R^{\tau \times \tau}_+, \theta \geq 0} \quad &\sum_i \Big[ (d_i^2 - \theta) \cdot z_i + 1 \Big]_+ \label{eq:assoc_lp} \\
\text{s.t.} \quad &d_i^2 = \sum_{t=1}^\tau \sum_{t'=1}^\tau \alpha_{t, t'} \cdot \lVert \bm{\Phi}_t \cdot \vec h_i - \bm{\Phi}_{t'} \cdot \vec m_i \rVert^2 & \forall i \notag
\end{align}
where $[x]_+ = \max\{0, x\}$ denotes the hinge loss. Note that this loss is zero if and only if
$d_i^2 - \theta < -1$ if $z_i = +1$ and $d_i^2 - \theta > 1$ if $z_i = -1$, i.e.\ if and only if
all associations are correct and a margin of safety for the associations is maintained.
The $d_i^2$ term in this problem further corresponds exactly to the squared distance
$\lVert \phi(\vec h_i) - \psi(\vec m_i)\rVert^2$ by setting $\phi(\vec h_i) = (\sqrt{\alpha_{1, 1}} \cdot \bm{\Phi}_1, \ldots, \sqrt{\alpha_{1, \tau}} \cdot \bm{\Phi}_1, \ldots, \sqrt{\alpha_{\tau, \tau}} \cdot \bm{\Phi}_\tau) \cdot \vec h_i$
and $\psi(\vec m_i) = (\sqrt{\alpha_{1, 1}} \cdot \bm{\Phi}_1, \ldots, \sqrt{\alpha_{1, \tau}} \cdot \bm{\Phi}_\tau, \ldots, \sqrt{\alpha_{\tau, \tau}} \cdot \bm{\Phi}_\tau) \cdot \vec m_i$,
where the commas indicate row-wise concatenation and $\alpha_{i, j}$ is the entry of the $i$th row
and $j$th column of $\bm{A}$. This is a highly sparse linear program, which can thus be solved efficiently with standard LP solvers.
We emphasize that the same scheme works for pre-defined non-linear
operators, similar to multiple kernel learning \cite{MKL}.
Further note that we can omit operators in the concatenation where $\alpha_{t, t'} = 0$, which we can
incentivize with an L1 regularization term.
This completes the construction of the associative RMM
$(\bm{U}, \bm{W}, \vec b, \sigma, \vec h_0, Q, \cls, \phi, \psi, \theta, \out)$.
The full training algorithm for an associative RMM is shown in Algorithm~\ref{alg:train_armm}.

\section{Experiments and Results}

In this section, we introduce our seven benchmark tasks, strategies for generating the required memory
addresses, explain the experimental setup, and present our results.

\subsection{Benchmark Tasks}

\begin{figure*}
\begin{center}
\begin{tikzpicture}
\tikzstyle{copy_region}=[orange3, pattern=north east lines, pattern color=orange3, semithick, opacity=0.7]

\begin{groupplot}[view={0}{90}, xlabel={$t$}, ymin=0, ymax=8,
group style={group size=2 by 3,
	x descriptions at=edge bottom,y descriptions at=edge left,
	horizontal sep=0.4cm, vertical sep=0.2cm},
width=4cm, height=3cm,
colormap={tango}{color(0cm)=(white); color(1cm)=(skyblue3)}]
\nextgroupplot[title={repeat copy$\strut$},ymax=9,ylabel={input$\strut$}]
\addplot3[surf,mesh/ordering=y varies,mesh/rows=10,shader=flat corner] file {repeat_copy_input.csv};
\draw[copy_region] (axis cs:1,0.05) rectangle (axis cs:4,8);
\node[inner sep=0] (rc_block) at (axis cs:2.5,0) {};
\nextgroupplot[title={associative recall$\strut$},ymax=7,height=2.7cm, yshift={0.15cm}] 
\addplot3[surf,mesh/ordering=y varies,mesh/rows=8,shader=flat corner] file {associative_recall_input.csv};
\draw[copy_region, pattern=crosshatch dots] (axis cs:0,0.05) rectangle (axis cs:3,5.95);
\draw[copy_region, pattern=crosshatch dots] (axis cs:10,0.05) rectangle (axis cs:13,5.95);
\draw[copy_region] (axis cs:3,0.05) rectangle (axis cs:6,5.95);
\node (ar_block1) at (axis cs:1.5,6) {};
\node (ar_block1_repeat) at (axis cs:11.5,6) {};
\node (ar_block2) at (axis cs:4.5,0) {};
\nextgroupplot[ymax=1,height=1.7cm,ylabel={address$\strut$}, yticklabels={,,}, ylabel style={yshift=-0.05cm}]
\addplot3[surf,mesh/ordering=y varies,mesh/rows=2,shader=flat corner] file {repeat_copy_states.csv};
\nextgroupplot[ymax=1,height=1.7cm,yticklabels={,,}, yshift={-0.3cm}]
\addplot3[surf,mesh/ordering=y varies,mesh/rows=2,shader=flat corner] file {associative_recall_states.csv};
\nextgroupplot[ymax=8,ylabel={output$\strut$},height=2.9cm]
\addplot3[surf,mesh/ordering=y varies,mesh/rows=9,shader=flat corner] file {repeat_copy_output.csv};
\draw[copy_region] (axis cs:1,0.05) rectangle (axis cs:4,8);
\draw[copy_region] (axis cs:5,0.05) rectangle (axis cs:8,8);
\draw[copy_region] (axis cs:9,0.05) rectangle (axis cs:12,8);
\draw[copy_region] (axis cs:13,0.05) rectangle (axis cs:16,8);
\node[inner sep=0] (rc_block_copy1) at (axis cs:2.5,8) {};
\node[inner sep=0] (rc_block_copy2) at (axis cs:6.5,8) {};
\node[inner sep=0] (rc_block_copy3) at (axis cs:10.5,8) {};
\node[inner sep=0] (rc_block_copy4) at (axis cs:14.5,8) {};
\nextgroupplot[ymax=6,height=2.6cm,yshift={-0.3cm}]
\addplot3[surf,mesh/ordering=y varies,mesh/rows=7,shader=flat corner] file {associative_recall_output.csv};
\draw[copy_region] (axis cs:13,0.05) rectangle (axis cs:15.95,5.95);
\node (ar_block2_recall) at (axis cs:14.5,6) {};
\end{groupplot}

\begin{scope}[shift={(6.5,0.2)}]
\begin{groupplot}[xlabel={$t$},xmin=0,xmax=1535,
group style={group size=1 by 3,
	x descriptions at=edge bottom,y descriptions at=edge left,
	horizontal sep=0.4cm, vertical sep=0.25cm},
width=8cm, height=2.8cm,
xtick={0,256,512,768,1024,1280}]
\nextgroupplot[title={\sigCopy$\strut$}, ytick={-3,0,3}]
\addplot[semithick, skyblue3] table[x=t,y=x1,col sep=tab] {signal_copy.csv};
\addplot[semithick, scarletred3]  table[x=t,y=x2,col sep=tab] {signal_copy.csv};
\draw[copy_region] (axis cs:0,-4) rectangle (axis cs:256,3.5);
\draw[copy_region, pattern=crosshatch dots] (axis cs:256,-4) rectangle (axis cs:512,3.5);
\node[inner sep=0] (sc_block1) at (axis cs:128,-4) {};
\node[inner sep=0] (sc_block2) at (axis cs:384,-4) {};

\nextgroupplot[height=1.9cm, ytick={0,1,2}, yticklabels={0,,}]
\addplot[semithick, skyblue3] table[x=t,y=q,col sep=tab] {signal_copy.csv};
\nextgroupplot[ytick={-3,0,3}]
\addplot[semithick, skyblue3] table[x=t,y=y,col sep=tab] {signal_copy.csv};
\draw[copy_region] (axis cs:768,-4)  rectangle (axis cs:1024,3.5);
\draw[copy_region, pattern=crosshatch dots] (axis cs:1024,-4) rectangle (axis cs:1280,3.5);
\draw[copy_region] (axis cs:1280,-4) rectangle (axis cs:1536,3.5);

\node[inner sep=0] (sc_block1_copy1) at (axis cs:896,3.5) {};
\node[inner sep=0] (sc_block2_copy1) at (axis cs:1152,3.5) {};
\node[inner sep=0] (sc_block1_copy2) at (axis cs:1408,3.5) {};
\end{groupplot}
\end{scope}

\begin{scope}[shift={(14,0)}]
\begin{groupplot}[view={0}{90}, xlabel={$t$}, ymin=0, ymax=28,
group style={group size=1 by 3,
	x descriptions at=edge bottom,y descriptions at=edge left,
	vertical sep=0.2cm},
width=4cm, height=3cm,
colormap={tango}{color(0cm)=(white); color(1cm)=(skyblue3)}]
\nextgroupplot[title={image recall$\strut$}]
\addplot3[surf,mesh/ordering=y varies,mesh/rows=29,shader=flat corner] file {image_copy_input.csv};
\draw[copy_region] (axis cs:1,0.05) rectangle (axis cs:27,28);
\node[inner sep=0] (ic_block) at (axis cs:14,0) {};
\nextgroupplot[ymax=1,height=1.7cm,yticklabels={,,}]
\addplot3[surf,mesh/ordering=y varies,mesh/rows=2,shader=flat corner] file {image_copy_states.csv};
\nextgroupplot[height=2.9cm]
\addplot3[surf,mesh/ordering=y varies,mesh/rows=29,shader=flat corner] file {image_copy_output.csv};
\draw[copy_region] (axis cs:28,0.05) rectangle (axis cs:55,28);
\draw[copy_region] (axis cs:56,0.05) rectangle (axis cs:83,28);
\draw[copy_region] (axis cs:84,0.05) rectangle (axis cs:111,28);
\node[inner sep=0] (ic_block_copy2) at (axis cs:42,28) {};
\node[inner sep=0] (ic_block_copy3) at (axis cs:70,28) {};
\node[inner sep=0] (ic_block_copy4) at (axis cs:98,28) {};
\end{groupplot}

\path[->, >=stealth, semithick, orange3, opacity=0.8]
(rc_block) edge[out=270, in=90] (rc_block_copy1)
(rc_block) edge[out=270, in=90] (rc_block_copy2)
(rc_block) edge[out=270, in=90] (rc_block_copy3)
(rc_block) edge[out=270, in=90] (rc_block_copy4)
(ar_block1) edge[out=90, in=90] (ar_block1_repeat)
(ar_block2) edge[out=270, in=90] (ar_block2_recall)
(sc_block1) edge[out=270, in=90, looseness=0.8] (sc_block1_copy1)
(sc_block1) edge[out=270, in=90, looseness=0.8] (sc_block1_copy2)
(sc_block2) edge[out=270, in=90, looseness=0.8] (sc_block2_copy1)
(ic_block) edge[out=270, in=90] (ic_block_copy2)
(ic_block) edge[out=270, in=90] (ic_block_copy3)
(ic_block) edge[out=270, in=90] (ic_block_copy4);
\end{scope}

\end{tikzpicture}
\vspace{-0.7cm}
\end{center}
\caption{An example from the repeat copy (left), associative recall (center left), \sigCopy\
(center right), and image recall (right) data sets,
each with input in the first, memory addresses in the second, and output in the third row.
Task-relevant blocks are highlighted with stripes/dots and arrows.}
\label{fig:data}
\end{figure*}
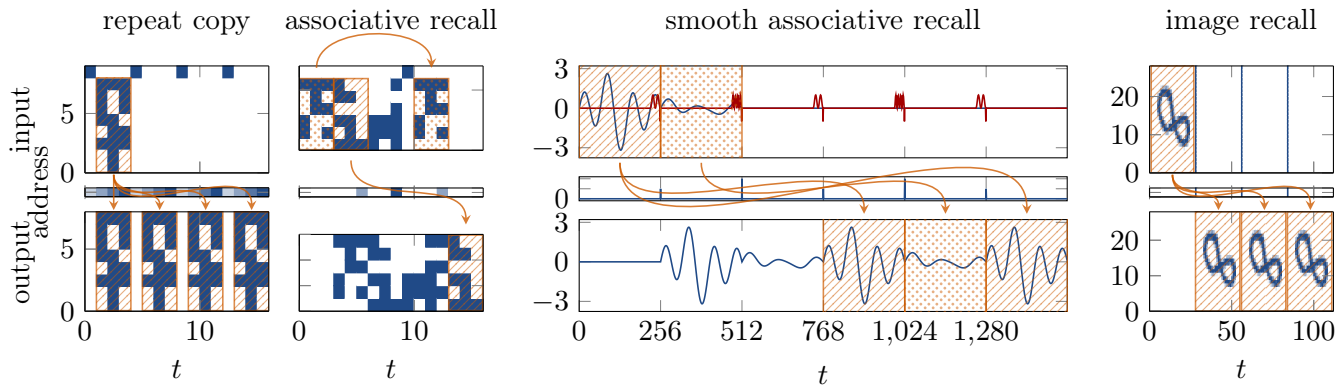

We evaluate reservoir memory machines (RMMs) on the following seven benchmark tasks:

\paragraph{latch \cite{RMM}} The input is a one-dimensional time series of length 9-200
that is always zero except for ones at 3 random time points. The desired output is zero
until the first one in the input, where it switches to one, then back to zero at the next one,
and back to one at the third one.
\paragraph{copy \cite{NTM_impl}} The input is a nine-dimensional time series of 1-20 random vectors from $\{0, 1\}^8$ on the first eight channels, followed by zeros. The last input channel is zero except
for a one right before and after the input vectors.
The desired output is a copy of the first eight channels as in Theorem~\ref{thm:copy}.
\paragraph{repeat copy \cite{NTM_impl}} The input is a nine-dimensional time series of
1-10 random vectors from $\{0, 1\}^8$ on the first eight channels, followed by zeros.
The desired output are $1-10$ copies of the first eight input channels. Each copy is preceded
by a one on the last input channel, which is zero otherwise. Refer to Figure~\ref{fig:data} (left)
for an example.
\paragraph{associative recall \cite{NTM_impl}} The input is a seven-dimensional time series with
2-6 random blocks á 3 random vectors from $\{0, 1\}^6$ on the first six channels, followed by
a one on the seventh channel and then a random repeated block from the previous input. The
desired output is the block \emph{after} the presented element. Refer to
Figure~\ref{fig:data} (center left) for an example.
\paragraph{\sigCopy} The input is a two-dimensional time series of two smooth random wavelets
of length $256$ on the first channel, followed by 1-10 blocks of $256$ zeros each.
On the second channel, each block of $256$ time steps ends with one of two marker wavelets of
length $32$. The first block with marker one, the second with marker two, and a random marker
(one or two) otherwise. The desired output responds to each marker with
its associated wavelet in the respective next $256$ time steps. Refer to Figure~\ref{fig:data}
(center right) for an example.
\paragraph{Image recall} The input is a random 28 x 28 grayscale image from the MNIST data set,
followed by $1-10$ vectors of ones. After each such vector, the output should be a copy of the
input image. Refer to Figure~\ref{fig:data} (right) for an example.
\paragraph{FSMs} For finite state machine learning we construct Moore machines with
$\inps = \outps = \{(1, 0)^T, (0, 1)^T\}$, $\states = \{1, 2, 3, 4\}$ and randomly sampled
transition as well as output functions. As training data, we construct all sequences with
exactly one repeated state in the Moore machine as suggested in Theorem~\ref{thm:fsm}.
The test data consists of much longer sequences of length $256$ over $\inps$ and the
output the Moore machine would predict.

We note that, among these tasks, only latch and FSM can be solved straightforwardly with
Moore machines. Copy, repeat copy, image recall, and associative recall require exponentially many states
by the same argument as in Theorem~\ref{thm:copy}, and \sigCopy\ is not solvable with a Moore
machine because there are infinitely many smooth wavelets that could be generated.

\subsection{Generating Memory Addresses}

To train RMMs, we require example memory address sequences for the training data.
The optimal strategy for such example address sequences depends on the task. We generally
recommend the following strategies:

For tasks with discrete and abstract outputs such as latch, we suggest to apply a clustering
on the target outputs. More specifically, for latch the memory address sequence becomes the training output $+1$ (i.e.\ naming the clusters 1 and 2).

For tasks which aim at storing information and recalling it in the same order, such as copy and repeat
copy, we suggest to use the state sequence $1, \ldots, T, 1, \ldots, T, \ldots, 1, \ldots, T$,
i.e.\ to enumerate each input vector and then recall each vector to generate the copy.
This is equivalent to the proof of Theorem~\ref{thm:copy}.

For block-wise recall tasks, such as \sigCopy, image recall, and associative recall we suggest to
store the state after each block in memory and load it when recalling the block. Otherwise, the memory
address should be zero.
More specifically, for \sigCopy\ the state sequence is zero
except after each block of $256$ where it is $1$ for marker one and $2$ for marker two.
For image recall, the address is $1$ at positions 29, 57, etc.\ and zero otherwise.
For associative recall the state sequence is $0, 0, 0, 0, 0, 1, 0, 0, 2, \ldots$, i.e.\ we
mark each block of 3, except the first one, and recall the index of the state to be recalled
after the presented item.

For FSM learning we use the ground-truth FSM to provide the state labels. We also evaluate the
alternative of first learning an FSM from data and using that to provide labels.

For repeat copy, associative recall, \sigCopy, and image recall, example inputs (top), outputs (bottom),
and address sequences (center) are shown in Figure~\ref{fig:data}.

\subsection{Experimental Setup}

We use a standard RMM with $64$ neurons for latch, \sigCopy, and FSM learning, a standard
RMM with $256$ neurons for copy and repeat copy, a standard RMM with $512$ neurons for image
recall, and an associative RMM with $256$ neurons for
associative recall. We compare against echo state networks \cite[ESN]{ESN} without an external memory
and the same number of neurons. We compare three kinds of reservoir, namely Gaussian random numbers
normalized to a spectral radius $< 1$ (rand), cycle reservoirs with jumps \cite[CRJ]{CRJ}, and Legendre delay networks \cite[LDN]{LMU}.
In each case we optimize the reservoir hyperparameters via random search with $20$ trials and
$3$ repeats per trial. An exception is the time horizon $T$ of a Legendre delay network because it follows
logically from the task, namely the time series length for latch,
$20$ for copy, $10$ for repeat copy, $6\cdot 3$ for associative recall, $256$ for smooth associative
recall, $28$ for image recall, and $4$ for FSMs.
For RMMs, we add the kernel of the SVM classifiers (linear or RBF with automatic bandwidth choice)
to the hyperparameters to be trained.

On the first four tasks, we additionally compare against deep learning models with the same number
of neurons, namely a gated recurrent unit \cite[GRU]{GRU}, and a deep version of our reservoir memory
machine with a GRU as a recurrent controller (GRU-MM), a softmax state classifier, and linear output. We
train this GRU-MM to minimize the mean squared error on the output plus the crossentropy loss on the
state predictions.
We train the deep models with an ADAM optimizer with learning rate $10^{-3}$, weight decay of
$10^{-8}$, and minibatch size of $32$. We stop the training after $1000$ minibatches or
if the loss is below $10^{-3}$.

All models are trained on $90$ training sequences and $10$ test sequences from the data sets.
We repeat all experiments $20$ times for the reservoir models and $3$ times for the deep models.
All experiments were performed on a desktop PC with Intel core i9-10900 CPU and 32 GB RAM.

All experimental source code and reference implementations are available at
\url{https://gitlab.com/bpaassen/rmm}.

\subsection{Results}

\begin{table*}
\centering
\caption{The root mean square error for all data sets and all models ($\pm$ std.\ dev.).}
\label{tab:results}
\begin{tiny}
\begin{tabular}{ccccccccc}
dataset & rand & CRJ & LDN & GRU & GRU-MM & rand-RMM & CRJ-RMM & LDN-RMM \\
\cmidrule(lr){1-1}\cmidrule(lr){2-4}\cmidrule(lr){5-6}\cmidrule(lr){7-9}
latch & $0.66 \pm 0.29$ & $0.50 \pm 0.04$ & $0.53 \pm 0.02$ & $0.05 \pm 0.04$ & $\bm{0.00 \pm 0.00}$ & $\bm{0.00 \pm 0.00}$ & $\bm{0.00 \pm 0.00}$ & $\bm{0.00 \pm 0.00}$\\
copy & $0.39 \pm 0.02$ & $0.45 \pm 0.01$ & $0.34 \pm 0.01$ & $0.39 \pm 0.01$ & $\bm{0.03 \pm 0.00}$ & $0.43 \pm 0.03$ & $0.48 \pm 0.02$ & $0.09 \pm 0.06$\\
repeat copy & $0.44 \pm 0.01$ & $0.46 \pm 0.01$ & $0.44 \pm 0.02$ & $0.45 \pm 0.02$ & $0.02 \pm 0.00$ & $0.37 \pm 0.02$ & $0.43 \pm 0.01$ & $\bm{0.01 \pm 0.02}$\\
smooth recall & $12.21 \pm 22.73$ & $12.07 \pm 22.76$ & $11.06 \pm 19.69$ & $11.49 \pm 11.90$ & $11.50 \pm 11.92$ & $12.21 \pm 22.72$ & $12.05 \pm 22.76$ & $\bm{\,\,\,4.79 \pm 20.21}$\\
image recall & $69.92 \pm 2.92\,\,\,$ & $88.60 \pm 47.37$ & $\,\,\,97.83 \pm 193.09$ & - & - & $70.22 \pm 3.29\,\,\,$ & $103.48 \pm 60.74\,\,\,$ & $\bm{26.91 \pm 9.59\,\,\,}$\\
FSMs & $0.41 \pm 0.20$ & $0.72 \pm 0.39$ & $0.56 \pm 0.17$ & - & - & $\bm{0.00 \pm 0.00}$ & $\bm{0.00 \pm 0.00}$ & $\bm{0.00 \pm 0.00}$\\
assoc.\ recall & $0.42 \pm 0.01$ & $0.31 \pm 0.01$ & $0.31 \pm 0.01$ & - & - & $0.43 \pm 0.02$ & $0.20 \pm 0.05$ & $\bm{0.10 \pm 0.08}$\\
\end{tabular}
\end{tiny}
\end{table*}

The average root mean square errors ($\pm$ std.) for all models on all data sets is shown in
Table~\ref{tab:results}. As can be seen, all RMM variants outperform ESNs on latch and FSM learning,
but only the LDN-RMM variant also achieves better results on copy, repeat copy, associative
recall, smooth associative recall, and image recall.
This is likely due to the fact that LDNs guarantee lossless reconstruction of past inputs,
which is required for these five tasks.
In comparison to the deep learning models we note that the standard GRU performs well on latch,
but comparable to standard ESNs on copy, repeat copy, and \sigCopy, which replicates earlier
results on differentiable neural computers \cite{NTM}. By contrast, a GRU-MM can solve all tasks
but \sigCopy, but the error remains comparable to the LDN-RMM.

Regarding FSM learning, we also consider a setting where we first learn a finite state
machine via Gold's algorithm \cite{FSM_learning} on the training sequences and then use
the states of this surrogate FSM as training memory address sequence.
This retains a perfect result in $19$ out of $20$ repeats but has one repeat with $0.32$
error, yielding an error of $0.02 \pm 0.07$ for all RMM variants, which is still close to
optimal.

\begin{table*}
\centering
\caption{The average runtime (without hyperparameter optimization) in seconds for all data sets and all models ($\pm$ std.\ dev.).}
\label{tab:runtimes}
\begin{tiny}
\begin{tabular}{ccccccccc}
dataset & rand & CRJ & LDN & GRU & GRU-MM & rand-RMM & CRJ-RMM & LDN-RMM \\
\cmidrule(lr){1-1}\cmidrule(lr){2-4}\cmidrule(lr){5-6}\cmidrule(lr){7-9}
latch & $0.09 \pm 0.00$ & $\bm{0.09 \pm 0.00}$ & $0.11 \pm 0.00$ & $510.27 \pm 8.95$ & $1180.03 \pm 17.25$ & $0.19 \pm 0.02$ & $0.18 \pm 0.02$ & $0.20 \pm 0.02$\\
copy & $0.09 \pm 0.01$ & $\bm{0.03 \pm 0.00}$ & $0.04 \pm 0.00$ & $448.03 \pm 6.22$ & $475.95 \pm 8.50$ & $1.83 \pm 0.17$ & $2.78 \pm 0.12$ & $0.65 \pm 0.03$\\
repeat copy & $0.23 \pm 0.04$ & $\bm{0.06 \pm 0.01}$ & $0.07 \pm 0.01$ & $1063.18 \pm 29.97$ & $\,\,\,869.85 \pm 20.84$ & $1.88 \pm 0.25$ & $6.20 \pm 0.63$ & $1.81 \pm 0.27$\\
smooth recall & $1.91 \pm 0.05$ & $\bm{1.77 \pm 0.04}$ & $1.98 \pm 0.04$ & $12147.15 \pm 87.73\,\,\,$ & $27377.17 \pm 102.16$ & $2.96 \pm 0.08$ & $2.73 \pm 0.06$ & $2.93 \pm 0.08$\\
image recall & $1.81 \pm 0.13$ & $0.53 \pm 0.02$ & $\bm{0.50 \pm 0.03}$ & - & - & $18.63 \pm 0.66\,\,\,$ & $1.77 \pm 0.11$ & $28.84 \pm 32.88$\\
FSMs & $0.06 \pm 0.01$ & $\bm{0.04 \pm 0.01}$ & $0.06 \pm 0.01$ & - & - & $0.26 \pm 0.01$ & $0.23 \pm 0.01$ & $0.25 \pm 0.02$\\
assoc.\ recall & $0.10 \pm 0.01$ & $\bm{0.03 \pm 0.01}$ & $0.05 \pm 0.02$ & - & - & $1.39 \pm 0.14$ & $1.70 \pm 0.27$ & $1.37 \pm 0.17$\\
\end{tabular}
\end{tiny}
\end{table*}

Table~\ref{tab:runtimes} shows the time needed for training and prediction (without hyperparameter
optimization; averaged across experimental repeats) as measured by Pythons
\texttt{time} function. Unsurprisingly, we observe that CRJ-ESNs are the fastest because
they operate with a highly sparse matrix that is fast to initialize.
We also note that LND-RMMs are roughly 20 times slower than LND-ESNs but remain in
the second range. The overhead is driven by the SVM training, which
is less costly on data sets with small explicit memory and, thus, fewer classes
(such as latch and \sigCopy).
By contrast, GRU-MMs are roughly $1000$ times slower
compared to LND-RMMs and take minutes to hours to train.

\section{Conclusion}

We introduced a novel reservoir memory machine (RMM) architecture which extends echo state networks
with an external memory and thereby becomes computationally strictly more powerful than finite state
machines, whereas contractive echo state networks cannot recognize some regular languages.
In addition to these theoretical results, we showed that RMMs can solve several benchmark tasks of 
differentiable neural computers which are out of reach for standard recurrent neural networks.
These successes require a sacrifice, namely that examples of memory access behavior need to be supplied
as part of the training data.
Future research should investigate how optimal memory access behavior can instead be learned from data,
perhaps via heuristics. Still, our model makes neural computation much more efficient to train, requiring
both less time and less training data compared to differentiable neural computers.

\section*{Acknowledgment}

Funding by the German Research Foundation (DFG) under grant numbers PA 3460/1-1 and PA 3460/2-1
is gratefully acknowledged.

\bibliographystyle{plainnat}
\bibliography{literature}

\clearpage

\begin{appendix}
\section{Conversion of FSMs to RNNs}

As an appendix to our main paper, we provide here a variant of a proof that any finite state machine
can be simulated with a recurrent neural net, using neurons as representations of the states. Note that
this is opposed to our simulation via reservoir memory machines, where we use the entire state vector,
and not just a single neuron, as representation of a Moore machine state.

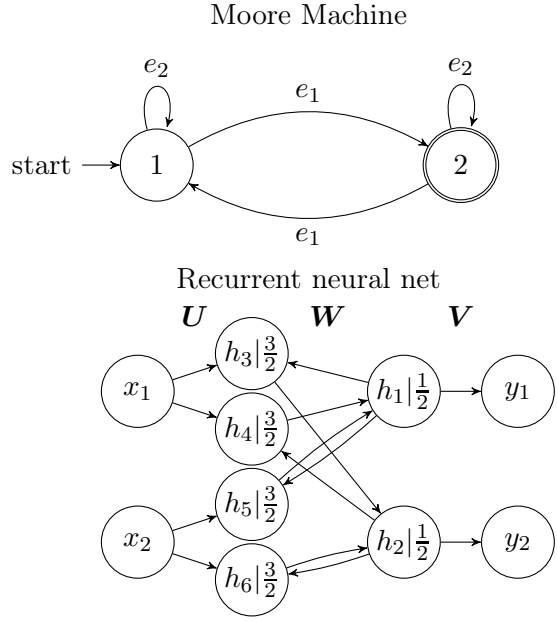
\begin{figure}
\begin{center}
\begin{tikzpicture}[>=stealth']

\begin{scope}
\node at (0,2) {Moore Machine};

\node[state,initial]   (q0) at (-2,0) {$1$};
\node[state,accepting] (q1) at (+2,0) {$2$};

\path[->]
(q0) edge[bend left]  node[above] {$e_1$} (q1)
(q0) edge[loop above] node[above] {$e_2$} (q0)
(q1) edge[bend left]  node[below] {$e_1$} (q0)
(q1) edge[loop above] node[above] {$e_2$} (q1);

\end{scope}

\begin{scope}[shift={(-2.25,-4)}]
\node at (2.25,2.5) {Recurrent neural net};

\node[state] (x1) at (0,+1) {$x_1$};
\node[state] (x2) at (0,-1) {$x_2$};

\node[state, inner sep=0pt] (h1) at (3.5,+1) {$h_1|\frac{1}{2}$};
\node[state, inner sep=0pt] (h2) at (3.5,-1) {$h_2|\frac{1}{2}$};
\node[state, inner sep=0pt] (h3) at (1.5,+1.5) {$h_3|\frac{3}{2}$};
\node[state, inner sep=0pt] (h4) at (1.5,+0.5) {$h_4|\frac{3}{2}$};
\node[state, inner sep=0pt] (h5) at (1.5,-0.5) {$h_5|\frac{3}{2}$};
\node[state, inner sep=0pt] (h6) at (1.5,-1.5) {$h_6|\frac{3}{2}$};

\node[state] (y1) at (5,+1) {$y_1$};
\node[state] (y2) at (5,-1) {$y_2$};

\node (U) at (0.75,2) {$\bm{U}$};
\node (W) at (2.5,2)  {$\bm{W}$};
\node (V) at (4.25,2)    {$\bm{V}$};

\path[->]
(x1) edge (h3) edge (h4)
(x2) edge (h5) edge (h6)
(h1) edge (h3) edge[bend left=5] (h5) edge (y1)
(h2) edge (h4) edge[bend left=5] (h6) edge (y2)
(h3) edge (h2)
(h4) edge (h1)
(h5) edge[bend left=5] (h1)
(h6) edge[bend left=5] (h2);

\end{scope}

\end{tikzpicture}
\end{center}
\caption{An example for the translation from a Moore machine (top) to a recurrent neural network
(bottom). For simplicity, we do not show weight values because all weights are $1$.
The negative bias value for each neuron is denoted after a vertical bar.}
\label{fig:fsm_to_rnn}
\end{figure}

\begin{thm}\label{thm:fsm_to_rnn}
Let $(\states, \inps, \outps, \delta, \rho, q_0)$ be a Moore machine 
with $\states = \{1, \ldots, L\}$, $\inps = \{e_1, \ldots, e_m\}$ and $\outps = \{e_1, \ldots, e_K\}$,
where $e_i$ is the $i$th unit basis vector. Further, let $x_1, \ldots, x_T$ be any sequence over $\inps$.
Then, there exists a recurrent neural network $(\bm{U}, \bm{W}, \bm{V}, \vec b, \sigma, \vec h_0)$ with
$\bm{U} \in \R^{n \times m}$, $\bm{W} \in \R^{n \times n}$, $\bm{V} \in \R^{K \times n}$, $\vec b, \vec h \in \R^n$ and $n = L \cdot (m+1)$.

Further, let $\vec h_t$ be the state and $\vec y_t$ be the output of the recurrent neural net at step $t$
of processing the input sequence $x_1, \vec 0, x_2, \ldots, \vec 0, x_T$ according to Equation~\ref{eq:rnn},
and let $q_t$ be the state and $y_t$ be the output of the Moore machine at step $t$ of processing
the input sequence $x_1, \ldots, x_T$. Then it holds for all $t \in \{1, \ldots, T\}$:
$\vec h_{2t} = e_{q_t}$ and $\vec y_{2t} = y_t$.

\begin{proof}
The idea of our proof is quite simple. Our aim is to set up $\vec h_{2t-1}$ such that
the coordinates $L+1, \ldots, (m+1) \cdot L$ represent the tuple $(q_{t-1}, x_t)$ via one-hot coding,
which then makes it trivial to map $\vec h_{2t-1}$ to $\vec h_{2t} = e_{q_t}$.
	
In particular, we set the weight matrices $\bm{U}$, $\bm{W}$, and $\bm{V}$ as well as the
bias vector $\vec b$ as follows.
\begin{align*}
w_{k, l} &=
\begin{cases}
1 & \text{if } \lfloor (l-1) / L \rfloor = i, \mod(l, L) = j \text{ and } \delta(e_i, j) = k\\
1 & \text{if } k > L \text{ and } \mod(k, L) = l\\
0 & \text{otherwise}
\end{cases} \\
u_{k, i} &=
\begin{cases}
1 & \text{if } 0 < \lfloor (k-1) / L \rfloor = i \leq m \\
0 & \text{otherwise}
\end{cases} \\
b_k &= 
\begin{cases}
-\frac{1}{2} & \text{if } k \leq L\\
-\frac{3}{2} & \text{otherwise}
\end{cases} \\
v_{j, k} &=
\begin{cases}
1 & \text{if } \rho(k) = e_j \\
0 & \text{otherwise}
\end{cases}
\end{align*}
We finally set $\sigma$ as the Heaviside function, i.e.\ $\sigma(x) = 1$ if $x > 0$ and $\sigma(x) = 0$
otherwise, and $\vec h_0$ as $e_{q_0}$. An example of this translation is visualized in Figure~\ref{fig:fsm_to_rnn}
for the Moore machine $(\{1, 2\}, \{e_1, e_2\}, \{e_1, e_2\}, \delta, \rho, 0)$ with $\delta(e_1, 1) = \delta(e_2, 2) = 2$
and $\delta(e_1, 2) = \delta(e_2, 1) = 1$ as well as $\rho(1) = e_1$ and $\rho(2) = e_2$.
An example processing of the input sequence $e_1, e_2, e_2, e_1$ would be processed by
the neural net as shown in Table~\ref{tab:fsm_to_rnn}.

Now, consider the claim $\vec h_{2t} = e_{q_t}$. We prove this claim via induction over $t$.
First, for $t = 0$, the claim holds because we set $\vec h_0 = e_{q_0}$.
Next, consider $t > 0$ and let's inspect the state $\vec h_{2t-1}$.
We wish to show that $h_{2t-1, k} = 1$ if $e_{\lfloor (k-1) / L \rfloor} = x_t$ and
$\mod(k, L) = q_{t-1}$ and $h_{2t-1, k} = 0$ otherwise. Due to induction we know that
$\vec h_{2t-2} = e_{q_{t-1}}$. Hence, we know that the sum $\sum_l w_{k, l} \cdot h_{2t-2, l}$
is $1$ if $k > L$ as well as $\mod(k, L) = q_{t-1}$. In any other case, this sum is zero
because all coordinates $h_{2t-2, l}$ for $l > L$ are zero.

Further, we know that the sum $\sum_i u_{k, i} \cdot x_{t, i}$ is $1$ if there exists
an $i \in \{1, \ldots, m\}$ such that $\lfloor (k-1) / L \rfloor = i$ and $e_i = x_t$.
Otherwise, the sum is zero. Plugging these results together we obtain that the sum
$\sum_l w_{k, l} \cdot h_{2t-2, l} + \sum_i u_{k, i} \cdot x_{t, i}$ is zero if $k \leq L$,
is $1$ if $k > L$ and either $\mod(k, L) = q_{t-1}$ or $\lfloor (k-1) / L \rfloor = i$,
but not both, and is $2$ if both $\mod(k, L) = q_{t-1}$ and $\lfloor (k-1) / L \rfloor = i$.
Combining this with the definition of $b_k$ and the Heaviside function, we obtain
indeed that $h_{2t-1, k} = 1$ if $e_{\lfloor (k-1) / L \rfloor} = x_t$ and
$\mod(k, L) = q_{t-1}$ and $h_{2t-1, k} = 0$ otherwise.

Now, consider the coordinate $h_{2t, q_t}$. This coordinate must be one because the 
term $w_{q_t, L \cdot i + q_{t-1}} \cdot h_{2t-1, L \cdot i + q_{t-1}}$ for $e_i = x_t$
is $1$ and, hence, the sum $\sum_l w_{q_t, l} \cdot h_{2t-1, l} - \frac{1}{2}$ is positive.
For all other coordinates $k$, note that the conditions
$\lfloor (l-1) / L \rfloor = i, \mod(l, L) = j \text{ and } \delta(e_i, j) = k$ do not hold
and that all coordinates $h_{2t-1, l}$ for $l \leq L$ are zero. Hence, the sum
$\sum_l w_{k, l} \cdot h_{2t-1, l} - b_k$ for all $k \neq q_t$ is negative.
This concludes the proof by induction.

Once this result is established, showing that $\vec y_{2t} = y_t$ is straightforward.
We simply notice that $y_t = \rho(q_t)$ and $\vec y_{2t} = \bm{V} \cdot \vec h_{2t}
= \bm{V} \cdot e_{q_t}$, such that the sum $\sum_k v_{j, k} \cdot h_{2t, k}$ is $1$
if $k = q_t$ and $\rho(k) = e_j$ and zero otherwise, which concludes the proof.´
\end{proof}
\end{thm}

\begin{table*}
\caption{The processing of the example sequence $e_1, e_2, e_2, e_1$ via the recurrent neural
network shown in Figure~\ref{fig:fsm_to_rnn}. Note that the input symbol at time $t$ is
put into the network at time $2t$.}
\label{tab:fsm_to_rnn}
\begin{center}
\begin{tabular}{cccccccccc}
$t$ & $0$ & $1$ & $2$ & $3$ & $4$ & $5$ & $6$ & $7$& $8$ \\
\cmidrule(lr){1-1} \cmidrule(lr){2-10}
$\vec x_t$ & - &
$\begin{pmatrix} 1 \\ 0 \end{pmatrix}$ &
$\begin{pmatrix} 0 \\ 0 \end{pmatrix}$ &
$\begin{pmatrix} 0 \\ 1 \end{pmatrix}$ &
$\begin{pmatrix} 0 \\ 0 \end{pmatrix}$ &
$\begin{pmatrix} 0 \\ 1 \end{pmatrix}$ &
$\begin{pmatrix} 0 \\ 0 \end{pmatrix}$ &
$\begin{pmatrix} 1 \\ 0 \end{pmatrix}$ &
$\begin{pmatrix} 0 \\ 0 \end{pmatrix}$ \\
$\vec h_t$ &
$\begin{pmatrix} 1 \\ 0 \\ 0 \\ 0 \\ 0 \\ 0 \end{pmatrix}$ &
$\begin{pmatrix} 0 \\ 0 \\ 1 \\ 0 \\ 0 \\ 0 \end{pmatrix}$ &
$\begin{pmatrix} 0 \\ 1 \\ 0 \\ 0 \\ 0 \\ 0 \end{pmatrix}$ &
$\begin{pmatrix} 0 \\ 0 \\ 0 \\ 0 \\ 0 \\ 1 \end{pmatrix}$ &
$\begin{pmatrix} 0 \\ 1 \\ 0 \\ 0 \\ 0 \\ 0 \end{pmatrix}$ &
$\begin{pmatrix} 0 \\ 0 \\ 0 \\ 0 \\ 0 \\ 1 \end{pmatrix}$ &
$\begin{pmatrix} 0 \\ 1 \\ 0 \\ 0 \\ 0 \\ 0 \end{pmatrix}$ &
$\begin{pmatrix} 0 \\ 0 \\ 0 \\ 1 \\ 0 \\ 0 \end{pmatrix}$ &
$\begin{pmatrix} 1 \\ 0 \\ 0 \\ 0 \\ 0 \\ 0 \end{pmatrix}$ \\
$\vec y_t$ & - &
$\begin{pmatrix} 0 \\ 0 \end{pmatrix}$ &
$\begin{pmatrix} 1 \\ 0 \end{pmatrix}$ &
$\begin{pmatrix} 0 \\ 0 \end{pmatrix}$ &
$\begin{pmatrix} 0 \\ 1 \end{pmatrix}$ &
$\begin{pmatrix} 0 \\ 0 \end{pmatrix}$ &
$\begin{pmatrix} 0 \\ 1 \end{pmatrix}$ &
$\begin{pmatrix} 0 \\ 0 \end{pmatrix}$ &
$\begin{pmatrix} 1 \\ 0 \end{pmatrix}$
\end{tabular}
\end{center}
\end{table*}

\end{appendix}

\end{document}